%% file: arxiv.tex
\documentclass[11pt, a4paper, logo, copyright]{googledeepmind}
\usepackage[round]{natbib}

\usepackage{microtype}
\usepackage{graphicx}
\usepackage{subfigure}
\usepackage{booktabs} 
\usepackage{hyperref}

\usepackage{amsmath}
\usepackage{amssymb}
\usepackage{mathtools}
\usepackage{amsthm}

\usepackage[capitalize,noabbrev]{cleveref}

\theoremstyle{plain}

\theoremstyle{definition}

\theoremstyle{remark}

\DeclareMathOperator*{\obs}{obs}
\DeclareMathOperator*{\act}{act}
\DeclareMathOperator*{\enc}{enc}
\DeclareMathOperator*{\bott}{bot}

\DeclareMathOperator*{\pos}{pos}
\DeclareMathOperator*{\sg}{sg}
\DeclareMathOperator*{\argmin}{argmin}
\DeclareMathOperator*{\argmax}{argmax}

\title{Learning Cognitive Maps from Transformer Representations for Efficient Planning in Partially Observed Environments}
\author[1]{Antoine Dedieu}
\author[1]{Wolfgang Lehrach}
\author[1]{Guangyao Zhou}
\author[1]{Dileep George}
\author[1]{Miguel L\'{a}zaro-Gredilla}
\affil[1]{Google DeepMind}
\correspondingauthor{adedieu@deepmind.com}

\begin{abstract}
Despite their stellar performance on a wide range of tasks, including in-context tasks only revealed during inference, vanilla transformers and variants trained for next-token predictions (a) do not learn an explicit world model of their environment which can be flexibly queried and (b) cannot be used for planning or navigation. In this paper, we consider partially observed environments (POEs), where 
an agent receives perceptually aliased observations as it navigates,
which makes path planning hard. We introduce a transformer with (multiple) discrete bottleneck(s), TDB, whose latent codes learn a compressed representation of the history of observations and actions. After training a TDB to predict the future observation(s) given the history, we extract interpretable cognitive maps of the environment from its active bottleneck(s) indices. These maps are then paired with an external solver to solve (constrained) path planning problems. First, we show that a TDB trained on POEs (a) retains the near-perfect predictive performance of a vanilla transformer or an LSTM while (b) solving shortest path problems exponentially faster. Second, a TDB extracts interpretable representations from text datasets, while reaching higher in-context accuracy than vanilla sequence models. Finally, in new POEs, a TDB (a) reaches near-perfect in-context accuracy, (b) learns accurate in-context cognitive maps (c) solves in-context path planning problems.
\end{abstract}

\begin{document}
\maketitle
\input{main_text_arxiv}

\end{document}

%% file: main_text_arxiv.tex

\section{Introduction}\label{sec:introduction}

Large vanilla transformers \citep{vaswani2017attention} trained for next-token prediction have been successfully applied to a wide range of applications such as natural language processing \citep{radford2018improving, chowdhery2022palm}, text-conditioned image generation \citep{ramesh2021zero}, reinforcement learning \citep {chen2021decision}, code writing \citep{chen2021evaluating}. These large language models (LLMs) also exhibit \textit{emergent} abilities \citep{wei2022emergent}, among which their capacity to \textit{in-context learn} (ICL), i.e., to adapt to a new task at inference time given a few examples \citep{brown2020language}. However, these models suffer from shortcomings that prevent them from being used for planning and navigation \citep{valmeekam2022large, mialon2023gaia}. In particular, they do not learn an interpretable world model \citep{momennejad2023evaluating} which can be flexibly queried.

Herein, we consider a suite of partially observed environments (POEs) \cite{chrisman1992reinforcement} where the agent receives perceptually aliased observations as it navigates; and cannot deterministically recover its spatial positions from its observations. Path planning in these POEs is hard: the planner has to disambiguate aliasing to locate itself, which requires modeling its history of observations and actions.

\begin{figure}[!t]
    \centering
    \includegraphics[width=.6\textwidth]{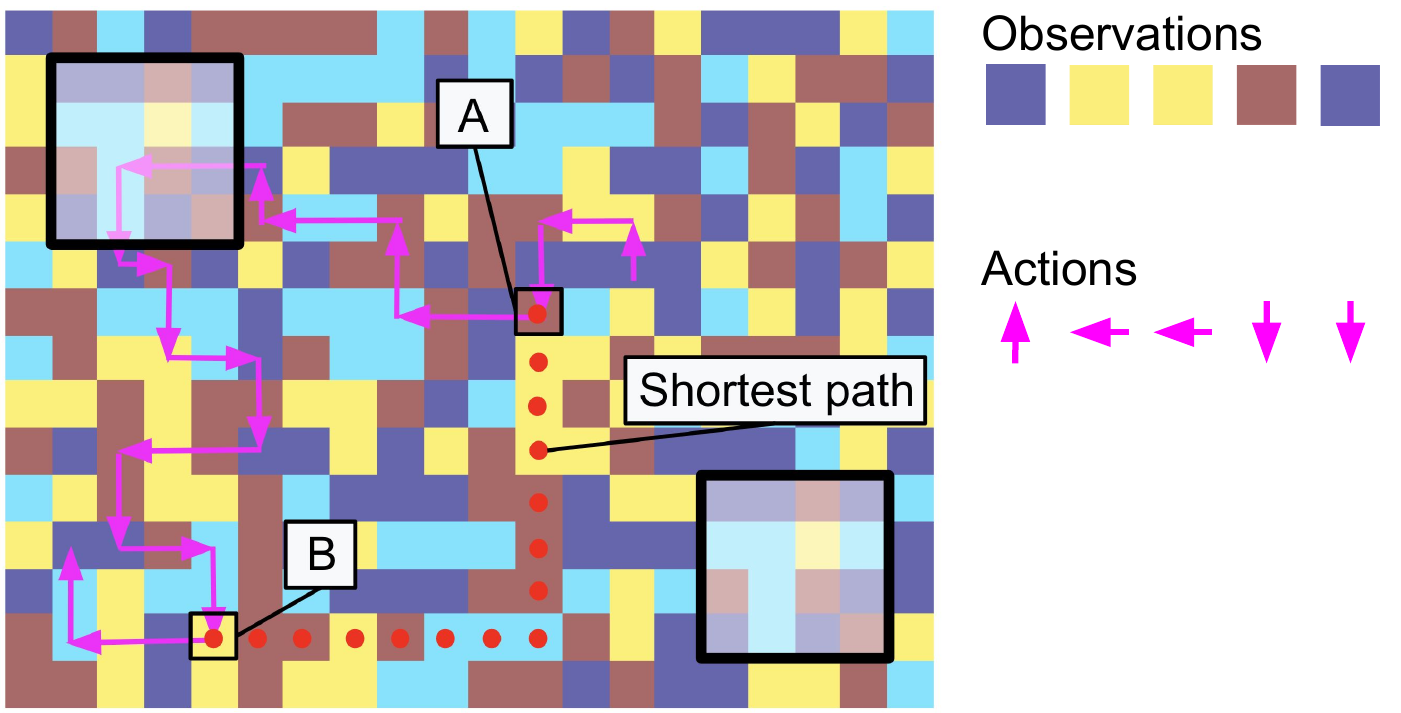}
    \caption{An agent is trained on random walks in an aliased room with no reward and unknown layout. At test time, given a novel random walk (in fuchsia), it has to find a shortest path (in red) between room positions A and B. While a vanilla transformer solves this path planning problem with forward rollouts, which can be exponentially expensive due to aliasing, our transformer variant pairs its learned cognitive map with an external solver.}
    \label{fig:motivation}
\end{figure}

\textbf{Example of a path planning problem in a POE that a vanilla transformer cannot solve: } We consider the partially observed $\small{15\times20}$ room in Fig.\ref{fig:motivation}, which only contains four unique observations. The room also has global aliasing: the $\small{4\times4}$ patch with a black border appears twice. An agent executes a series of discrete actions, each action leading to a discrete observation. The agent does not have access to any reward or to the room layout. At test time, it wants to find a shortest path (in red) between two room positions (\textbf{A} and \textbf{B}) it has been to. A transformer trained to predict the next observation given past observations and actions can only perform forward rollouts, which, due to aliasing, (a) scale exponentially in the $\ell_1$ distance between \textbf{A} and \textbf{B} (b) prevents it from even knowing when it reaches the target \textbf{B}.




In this paper, we propose the transformer with discrete bottleneck(s), TDB), which adds a single, or multiple, discrete bottleneck(s) \citep{van2017neural} on top of a transformer to compress the transformer outputs into a finite number of latent codes. 
We train TDB with an augmented objective, then extract a cognitive map from its active bottleneck(s) indices. First, we show that, on aliased rooms, aliased cubes with non-Euclidean dynamics and visually rich $3$D environments \citep{beattie2016deepmind}, these maps (a) disambiguate aliasing (b) nearly recover the ground truth dynamics and (c) can be paired with external solvers to solve path planning problems---like the one in Fig.\ref{fig:motivation}. Hence, \texttt{TDB} (a) retains the nearly perfect predictive abilities of vanilla transformers and LSTMs, (b) solves (constrained) path planning problems exponentially faster. Second, a \texttt{TDB} extracts an interpretable latent graph from a text dataset \citep{xie2021explanation} while achieving higher test accuracies than vanilla sequence models. Finally, when exploring a new POE at test time, \texttt{TDB} can (a) in-context predict the next observation given history (b) solve in-context path planning problems (c) learn accurate in-context cognitive maps.

The rest of this paper is organized as follows.  Sec.\ref{sec:related} discusses related work.  Sec.\ref{sec:model} details our proposed \texttt{TDB} model. Finally,  Sec.\ref{sec:exp} compares our method with vanilla transformer and LSTM in a variety of navigation and text experiments.

\section{Related work}\label{sec:related}


\textbf{Planning with LLMs: } Despite their success on simple benchmarks including maths \citep{cobbe2021training} and logics \citep{srivastava2022beyond}, growing evidence suggests that LLMs performance collapse on benchmarks that require stronger planning skills \citep{valmeekam2022large}. While \citet{pallagani2022plansformer} finetune LLMs for planning, \citet{guan2023leveraging} extract an approximate world model with an LLM, then further refine it with an external planner.

\textbf{Interpretable circuits: } Some works try to reverse-engineer LLMs to extract interpretable circuits, i.e., internal structures that drive certain model behaviors \citep{elhage2021mathematical, olsson2022context, nanda2023progress}. While circuit analysis can be performed at scale \citep{lieberum2023does}, it is labor intensive \citep{conmy2023towards}, 
and does not extract explicit structures to solve planning or navigation tasks. 

\textbf{Decision transformers: } Decision transformers \citep{chen2021decision} abstract reinforcement learning (RL) as sequence modeling: they can play Atari games and stack blocks with a robot arm \citep{reed2022generalist}. However, they only have been applied to shortest path problems on small graphs with $20$ nodes. Decision transformers have also access to a reward, which reveals information about the environment. They would struggle to solve the shortest path in Fig.\ref{fig:motivation}, where aliasing is high and no external reward is provided.

\textbf{Learning world models with a transformer: } \citet{lamb2022guaranteed} train a transformer with a multi-step objective on sequences of observations and actions, and successfully learned a \textit{minimal} world representation that either the agent can control or that affects its dynamics. However, the authors (a) consider fully visible settings and (b) learn representations by discarding the agent's history: their approach would not solve the shortest path in aliased settings as in Fig.\ref{fig:motivation}. In another approach, \citet{guo2022byol} learn future latent representations from current latent representations but do not explicitly extract interpretable world models that can be used for planning. In a different line of work, \citet{li2022emergent} show that transformers learn internal world models, that can be used to control some model's behavior. In contrast, here, we are interested in extracting a world model from a transformer that can be used to solve path planning problems exponentially faster than vanilla models.

\textbf{Clone-structured causal graphs (CSCGs): } 
CSCG \citep{george2021clone} is an interpretable probabilistic model that learns cognitive maps in aliased POEs and solves path planning tasks. 
However, CSCGs learn a block-sparse transition matrix over a large latent space via the expectation-maximization (EM) algorithm \citep{dempster1977maximum}. This large transition matrix occupies a large amount of memory and limits CSCGs' scalability---see Appendix \ref{sec:appendix_cscg_complexity} for a detailed discussion. 
CSCGs would also struggle to solve the ICL experiments in Sec.\ref{sec:exp_icl} without an external algorithm---see \citet{swaminathan2023schema}. In contrast, the transformer variant we introduce in this paper builds this same transition matrix sparsely, via local counts. Appendix \ref{sec:appendix_clone_structure} discusses a CSCG-inspired variant of our proposed model.


\section{A path planning-compatible transformer}\label{sec:model}

\subsection{Problem statement}

\textbf{Problem: }
We consider an agent executing a series of discrete actions $a_1,\ldots,a_{N-1}$ with $a_n\in\{1,\ldots, N_\text{actions}\}$, e.g. walking in a room. As a result of each action, the agent receives a perceptually aliased observation \cite{chrisman1992reinforcement}, resulting in the stream $x_1,\ldots,x_N$. The agent does not have access to any reward at any time. Our goal is to (a) predict the next observation given the history of past observations and actions and (b) build a world model that disambiguates aliased observations and that can be called by an external planner to solve path planning tasks.

\textbf{Trajectory representation: }
We represent a trajectory $\tau=(x_1, a_1, \ldots, a_{N-1}, x_{N})$ by alternating observations and actions. This representation allows us to consider the autoregressive objective of predicting the next observation given the history of past observations and actions:
\setlength{\abovedisplayskip}{10pt}
\begin{equation}\label{eq:transformer_objval}
\mathcal{L}_{\obs} = \sum_{n=1}^{N-1} \mathcal{L}_{\obs}(n) =-  \sum_{n=1}^{N-1} \log p(x_{n+1} | x_1, a_1, \ldots, x_n, a_n)
\setlength{\belowdisplayskip}{10pt}
\end{equation}

\subsection{Predicting the next observation with a transformer}\label{sec:vanilla_transformer}


We propose to train a vanilla transformer \citep{vaswani2017attention} to minimize Equation \eqref{eq:transformer_objval}. To do so, we first map each categorical observation $x_n$ (resp. action $a_n$) of $\tau$ to a linear embedding $E_{\obs}(x_n) \in \mathbb{R}^D$ (resp. $E_{\act}(a_n) \in \mathbb{R}^D$). Second, we feed the sequence of input embeddings  $z=\left(E_{\obs}(x_1), E_{\act}(a_1), \ldots, E_{\obs}(x_N),  E_{\act}(a_{N})\right)$ to a transformer with causal mask \citep{radford2018improving}, resulting in the output sequence $(T_1, \ldots, T_{2N})$. We only retain the outputs with \textit{even} indices $(T_2, T_4,\ldots, T_{2N})$: $T_{2n} \in \mathbb{R}^D$ is derived after applying the successive self-attention layers to the intermediate representations of $x_1, a_1, \ldots, x_n, a_{n}$. Finally, a linear mapping applied to $T_{2n}$ returns the predicted logits for the next observation $x_{n+1}$. We display this transformer architecture in Fig.\ref{fig:vanilla_transformer}, Appendix \ref{sec:appendix_vanilla_transformer}.

As we show in  Sec.\ref{sec:exp}, this transformer almost perfectly predicts the next observation on a suite of perceptually aliased POEs. However, it can only solve the path planning problem in Fig.\ref{fig:motivation} with forward rollouts, which have an exponential cost in the $\ell_1$ distance between the room positions \textbf{A} and \textbf{B}. 

\begin{figure*}[!t]
    \centering
    \includegraphics[width=.9\textwidth]{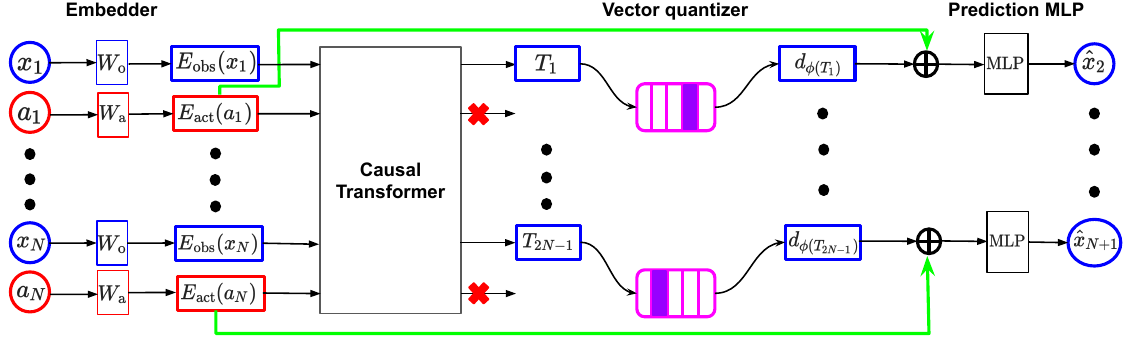}
    \caption{Our proposed transformer with a (single) discrete bottleneck. The respective linear embeddings of observations and actions go through a causal transformer. The observation outputs are compressed by the vector quantizer, then concatenated with the next action embedding in order to predict the next observation. Finally, a cognitive map of the environment is built from the active bottleneck indices.}
    \label{fig:transformer_discrete_bottleneck}
\end{figure*}

\subsection{A transformer with a discrete bottleneck}\label{sec:discrete_bottleneck}
To address these inherent limitations, we add a discrete bottleneck on top of a vanilla transformer, which compresses all the information needed to minimize Equation \eqref{eq:transformer_objval}.

After applying the causal transformer from Sec.\ref{sec:vanilla_transformer} to the sequence $z$ of observations and actions embeddings, we now only extract the encodings with \textit{odd} indices, which results in the stream $(T_1, T_3,\ldots, T_{2N-1})$: $T_{2n-1} \in \mathbb{R}^D$ is derived from $x_1, a_1, \ldots, a_{n-1}, x_n$, but is not aware of the action $a_n$. We then apply vector quantization \citep{van2017neural} and map each vector $T_{2n-1}, \forall n$ to the closest entry in a dictionary of latent codes $\mathcal{D} = (d_1, \ldots, d_K)$, where $d_k \in \mathbb{R}^D, ~\forall k$. That is, we introduce the operator
\setlength{\abovedisplayskip}{10pt}
\begin{equation}\label{eq:vector_quantize}
\phi(y) = \argmin_{k=1, \ldots, K} \| y - d_k \|_2^2, ~\forall y.
\end{equation}
\setlength{\belowdisplayskip}{10pt}
The quantized output of $T_{2n-1}$ is $d_{\phi(T_{2n-1})}$. Because $T_{2n-1}$ has not been exposed to $a_n$, we derive the next observations logits via $f\left(d_{\phi(T_{2n-1})} \oplus E_{\act}(a_n)\right)$, where $f$ is a two layers MLP and $\oplus$ represents the concatenation operator. We refer to our proposed transformer with a discrete bottleneck as \texttt{TDB}, and visualize it in Figure \ref{fig:transformer_discrete_bottleneck}.

\subsection{Augmenting the objective value}\label{sec:objective_augmentation}
The discrete bottleneck in  Sec.\ref{sec:discrete_bottleneck} introduces the additional objective term $\mathcal{L}_{\bott} = \sum_{n=1}^N \mathcal{L}_{\bott}(n)$ with
\setlength{\abovedisplayskip}{10pt}
\begin{equation}\label{eq:bottleneck_objval}
\mathcal{L}_{\bott}(n) = \| d_{\phi(e)} - \sg(e) \|_2^2 
+ \beta \| \sg\left(d_{\phi(e)} \right) - e\|_2^2,
\end{equation}
\setlength{\belowdisplayskip}{10pt}
where $e=T_{2n-1}$, $\beta=0.25$ and $\sg$ the stop-gradient operator \citep{van2017neural}. However, we show in Fig.\ref{fig:appendix_merging_single_step_aliased}, Appendix \ref{appendix:merging_single_step} that a TDB trained with the simple objective $\mathcal{L}_{\obs} + \mathcal{L}_{\bott}$ fails at disambiguating observations with identical neighbors. To allow the TDB to learn richer representations of the observations, we propose to augment the transformer objective via two solutions.

\textbf{Solution 1: Predicting the next $S$ observations.} At each timestep $n$, instead of only predicting the next observation---which corresponds to $S=1$---we predict the next $S$ observations and replace the loss $\mathcal{L}_{\obs}(n)$ in Equation \eqref{eq:transformer_objval} with
\setlength{\abovedisplayskip}{10pt}
\begin{equation}\label{eq:transformer_objval_ksteps}
\mathcal{L}^S_{\obs}(n) = - \frac{1}{S} \sum_{s=0}^{S-1} \log p(x_{n+s+1} | x_1, a_1, \ldots, x_n, a_n, \ldots, a_{n+s})
\end{equation}
\setlength{\belowdisplayskip}{10pt}
The model does not have access to any observation after $x_n$: it only sees the $S$ actions $a_n,\ldots, a_{n+S-1}$. The logits of the observation $x_{n+s+1}$ are predicted via $f^{(s)}\left(d_{\phi(T_{2n-1})} \oplus E_{\act}(a_n) \oplus \ldots \oplus E_{\act}(a_{n+s})\right)$, where $f^{(s)}$ is a two-layers MLP. In Sec.\ref{sec:exp}, we set $S=3$.

\textbf{Solution 2: Predicting the next encoding.}
Our second solution is to predict future latent representations \cite{guo2022byol}. We refer to the $\texttt{TDB}$ described so far as a student network, with weights $\Theta$, and introduce a teacher network \citep{grill2020bootstrap}, with the same architecture and whose weights $\Theta^{\text{teacher}}$ are an exponential moving average of $\Theta$. At each gradient update, we set $\Theta^{\text{teacher}} \leftarrow \alpha \Theta^{\text{teacher}} + (1-\alpha) \Theta$, where $\alpha$ is the decay rate, which we fix to $0.99$. 

The student network predicts both (a) the next observation and (b) the next teacher network's quantized encoding. This introduces the objective term $\mathcal{L}_{\enc} = \sum_{n=1}^{N-1} \mathcal{L}_{\enc}$ with 
\setlength{\abovedisplayskip}{10pt}
\begin{align}\label{eq:transformer_objval_byol}
&\mathcal{L}_{\enc}(n)=
\left\| 
\underbrace{\frac{g\left(d_{\phi(T_{2n-1})} \oplus E_{\act}(a_n)\right)}{\| g\left(d_{\phi(T_{2n-1})} \oplus E_{\act}(a_n)\right) \|_2 } }_{\text{using } \Theta}
-
\sg \left(\underbrace{\frac{d_{\phi(T_{2n + 1})}}{\| d_{\phi(T_{2n + 1})}\|_2}}_{\text{using } \Theta^{\text{teacher}}}\right)
\right\|_2^2 \nonumber
\end{align}
\setlength{\belowdisplayskip}{10pt}
where $g$ is a two-layer MLP with same hidden layer as $f$. 

\subsection{Extension to multiple discrete bottlenecks}\label{sec:mutiple_dbs}
\texttt{TDB} supports $\small{M>1}$ discrete bottlenecks. Each bottleneck induces its own (a) dictionary of latent codes $\mathcal{D}^{(i)}=(d^{(i)}_1, \ldots,d^{(i)}_K)$, (b) function $\phi^{(i)}$ which acts as in Equation \eqref{eq:vector_quantize} and (c) objective $\mathcal{L}_{\bott}^{(i)}$. The transformer output $T_{2n-1}$ passes through the discrete bottlenecks in parallel: the $i$th bottleneck returns the latent code $d^{(i)}_{ \phi^{(i)}(T_{2n-1}) }$. We get back to the previous case by defining (a) $d_{\phi(T_{2n-1})} = \oplus_{i=1}^M d^{(i)}_{ \phi^{(i)}(T_{2n-1}) }$ and (b) $\mathcal{L}_{\bott} = \sum_{i=1}^{M}\mathcal{L}_{\bott}^{(i)}$. As we show in  Sec.\ref{sec:exp}, multiple discrete bottlenecks make training faster and sometimes better. However, Appendix \ref{appendix:disentanglement} introduces a disentanglement metric which shows that they are highly redundant and do not learn disentangled representations.

\subsection{Learning a cognitive map from bottleneck indices}\label{sec:world_model}

When we use a single bottleneck, \texttt{TDB} maps each observation $x_n$ to the latent code index  $s_n=\phi(T_{2n-1})\in \{1, \ldots, K\}$. We define a count tensor $C$ which counts the empirical latent transitions: $C_{ijk} = \sum_{(s_n, a_n, s_{n+1})} \mathbf{1}(s_n=i, a_n=j, s_{n+1}=k)$. We then threshold $\mathcal{C}$ to build an action-augmented empirical transition graph $\mathcal{G}$. That is, $\mathcal{G}$ has an edge between the vertices $i$ and $k$ with the action $j^*=\argmax_{j} C_{ijk}$ iff. $C_{ij^*k} \ge t_{\text{ratio}} ~ c^*$, where $c^* = \max_{ijk} C_{ijk}$ and $t_{\text{ratio}}$ is a threshold. This edge indicates that the action $j^*$ leads from the latent node $i$ to the latent node $k$ a large number of times.

In practice (a) we use multiple discrete bottlenecks (see  Sec.\ref{sec:mutiple_dbs}) and cluster the bottleneck indices using the Hamming distance with a threshold $d_{\text{Hamming}}$ and (b) we map each discarded node of $\mathcal{C}$ to a retained node in $\mathcal{G}$ so that we can solve all the path planning problems in Sec.\ref{sec:exp}. We detail (a) and (b) in Appendix \ref{sec:appendix_cluster}. As a result, each node in $\mathcal{G}$ is paired with a collection of tuples of bottleneck indices. 

$\mathcal{G}$ is a cognitive map of the agent's environment which (a) is interpretable, (b) is action-conditioned, hence, models the agent's dynamics, and (c) can be paired with an external planner to solve path planning problems---see  Sec.\ref{sec:exp}. 

\subsection{Why does TDB learn an accurate map?}\label{sec:intuition}

The \texttt{TDB} latent index $\phi(T_{2n-1})$ associated with an observation $x_n$ given its history corresponds to a node in the latent graph $\mathcal{G}$ in Sec.\ref{sec:world_model}. For $\mathcal{G}$ to correctly model the environment's dynamics, when this node is active, the agent must always be at the same ground truth spatial position.

$d_{\phi(T_{2n-1})}$ does not know the next action $a_n$; and compresses all the information needed to minimize the loss at the $n$th step. When \texttt{TDB} predicts the next observation, $d_{\phi(T_{2n-1})}$ must then encode the next observation that each next possible action leads to. As we show in Appendix \ref{appendix:merging_single_step}, $d_{\phi(T_{2n-1})}$ may not disambiguate distinct spatial positions with identical neighbors. When \texttt{TDB} predicts the next $S$ observations (Solution 1,  Sec.\ref{sec:objective_augmentation}), $d_{\phi(T_{2n-1})}$ must now encode all the observations $x_{n+s}, 1\le s\le S,$ that each sequence of $S$ actions leads to. For a large $S$, the latent index $\phi(T_{2n-1})$ is encouraged to only be active at a unique spatial location. An alternative approach is to augment $d_{\phi(T_{2n-1})}$ so that it compresses all its neighbors' encoding (Solution 2).

\section{Computational results}\label{sec:exp} 

We assess the performance of our proposed \texttt{TDB} on four experiments: (a) synthetic $2$D aliased rooms and aliased cubes, (b) simulated $3$D environments, (c) text datasets and (d) mixtures of $2$D aliased rooms. Each experiment is run on a $2\times2$ grid of Tensor Processing Unit v2.

\subsection{Methods compared}\label{sec:methods_compared}

Each experiment compares the following methods:

$\bullet$ The causal transformer described in  Sec.\ref{sec:vanilla_transformer}, trained with the autoregressive objective in Equation \eqref{eq:transformer_objval}. The architecture considered has $4$ layers, $8$ attention heads, a context length of $400$, an embedding dimension of $256$, and an MLP hidden layer dimension of $512$. We use relative positional embedding \citep{dai2019transformer} and Gaussian error linear units activation functions \citep{hendrycks2016gaussian}.

$\bullet$ Several \texttt{TDB}s, using the same causal transformer. We refer to each model as \texttt{TDB}$(S, M)$ where (a) $S \in \{1, 3\}$ is the number of prediction steps (Solution 1,  Sec.\ref{sec:objective_augmentation}), (b) $M \in \{1, 4\}$ is the number of discrete bottlenecks (see  Sec.\ref{sec:mutiple_dbs})---each one contains $K=1000$ latent codes. We note \texttt{TDB}$(S, \enc, M)$ when we predict the next encoding (Solution 2,  Sec.\ref{sec:objective_augmentation}). After training, we build a cognitive map as in  Sec.\ref{sec:world_model} using $d_{\text{Hamming}}=0.25, t_{\text{ratio}}=0.1$.

$\bullet$ A vanilla LSTM \citep{hochreiter1997long} with hidden dimension of 256.

\subsection{Random walks in 2D aliased rooms}\label{sec:exp_aliased_room}

\textbf{Problem: }
We consider a $2$D aliased ground truth (GT) room of size $15\times20$, containing $\small{O=4}$ unique observations, and a $4\times4$ patch repeated twice---similar to Fig.\ref{fig:motivation}. An agent walking selects, at each timestep, a discrete action $a_n \in \{1, 2, 3, 4\}$ and collects the categorical observation $x_n \in \{1, \ldots, O\}$. The actions have unknown semantics: the agent does not know that $a_n=1$ corresponds to moving up. No assumptions about Euclidean geometry are made either. The training and test set contains $2048$ random walks of observations and actions of length $400$ each.

\medskip

\textbf{Training: }
We train each method in  Sec.\ref{sec:methods_compared} with Adam \citep{kingma2014adam} for $25$k training iterations, using a learning rate of $0.001$ and a batch size of $32$. For regularization, we use a dropout of $0.1$.

\medskip

\textbf{Solving path planning problems at test time: }
For a test sequence $x^{\text{test}}$ of length $N=400$, let $\pos_n$ be the GT \textit{unknown} $2$D spatial position of the observation $x_n$. At test time, given a context $C=50$, the path planning task is to derive a \textit{shortest path}, i.e. a sequence of actions, which leads from $\pos_C$ to $\pos_{N-C}$ in the GT room. We emphasize that (a) the model only observes $x_n$ and does not know $\pos_n$ (b) the path $(a_C,\ldots, a_{N-C-1})$ is a solution, referred to as \textit{fallback path}. 
Fig.\ref{fig:motivation} shows this problem for $N=40, C=5$.

For a vanilla sequence model, we autoregressively sample $100$ random rollouts of length $N-2C$ each and collect all the \textit{candidates}, i.e., the observations equal to $x^{\text{test}}_{N-C}$. Because of aliasing, the model cannot know whether a candidate is at the target position $\pos_{N-C}$. We use the tail of the last $C$ observations and actions and only return the candidates estimated to be at $\pos_{N-C}$. For a \texttt{TDB}, after learning the cognitive map $\mathcal{G}$, we map $x_C$ and $x_{N-C}$ to the bottleneck indices $\phi(T_{2C-1})$ and $\phi(T_{2(N-C)-1})$, then to two nodes of $\mathcal{G}$ using the Hamming distance. We then call the external solver \texttt{\small{networkx.shortest\_path}} \citep{hagberg2008exploring} to find the shortest path between these two nodes.  See Appendix \ref{appendix:planning_metric} for details about both procedures.

\paragraph{Metrics: }
We evaluate each method for four metrics.

$\bullet$ A \textit{prediction} metric, \texttt{TestAccu}: next observation accuracy on the test random walks. 

$\bullet$ Two \textit{path planning} metrics: we consider $200$ shortest paths problems and report (a) \texttt{ImpFallback}, the percentage of problems for which a model finds a \textit{valid} path which improves over the fallback path---a path (i.e. a sequence of actions) is valid if it correctly leads from $\pos_C$ to $\pos_{N-C}$ in the GT room---and (b) \texttt{RatioSP}, when a better valid path is found, the ratio between its length and an optimal path length---we derive an optimal path using the GT room.

$\bullet$ Normalized \textit{graph edit distance}, defined for two graphs $\mathcal{G}_1,\mathcal{G}_2$ as
\newline
$\text{NormGED}(\mathcal{G}_1, \mathcal{G}_2) = \text{GED}(\mathcal{G}_1, \mathcal{G}_2) ~~/~~ (\text{GED}(\mathcal{G}_1, \emptyset) + \text{GED}(\mathcal{G}_2, \emptyset))$ where \texttt{GED} is the graph edit distance \citep{sanfeliu1983distance} and $\emptyset$ is the empty graph. \texttt{NormGED} is lower than $1.0$ and is equal to $0.0$ iff. $\mathcal{G}_1$ and $\mathcal{G}_2$ are isomorphic. We use a timeout of $15$ minutes and empirical positions to accelerate the \texttt{GED} computation---see Appendix \ref{appendix:ged}.

\begin{table*}[!t]
\centering
\resizebox{0.9\textwidth}{!}{
\begin{tabular}{p{0.25\textwidth}p{0.15\textwidth} p{0.18\textwidth} p{0.15\textwidth} p{0.15\textwidth}}
    \toprule
    Method & TestAccu ($\%$) $\uparrow$ & ImpFallback ($\%$) $\uparrow$ & RatioSP $\downarrow$ & NormGED $\downarrow$ \\
    \toprule
    \toprule
    Vanilla transformer &  $99.00~(0.01)$  & $29.53~(0.75)$ & $16.82~(0.92)$ & ~~~~~~~~~~~~---\\
    \midrule
    Vanilla LSTM &  $98.85~(0.01)$  & $31.97~(0.66)$ & $16.81~(0.83)$ & ~~~~~~~~~~~~---\\
    \midrule
    \texttt{TDB}$(S=1, M=1)$ & $98.94~(0.05)$  & $6.61~(0.53)$ & $\mathbf{1.00}~(0.00)$ & $0.532~(0.005)$\\
    \midrule
    \texttt{TDB}$(S=1, \enc, M=1)$ &  $98.74~(0.04)$  & $99.20~(0.20)$ & $\mathbf{1.00}~(0.00)$ & $\mathbf{0.118}~(0.022)$\\
    \midrule
    \texttt{TDB}$(S=3, M=1)$ & $\mathbf{99.06}~(0.01)$  & $\mathbf{99.55}~(0.13)$ & $\mathbf{1.00}~(0.00)$ & $0.124~(0.019)$\\
    \midrule
    \texttt{TDB}$(S=3, M=4)$ & $99.00~(0.01)$  & $96.59~(0.15)$ & $\mathbf{1.00}~(0.00)$ & $0.186~(0.018)$\\
    \bottomrule
\end{tabular}}
\caption{Results averaged over $10$ aliased $15\times 20$ rooms with $O=4$ observations.  See main text for a description of the metrics. Arrows pointing up (down) indicate that higher (lower) is better. Our \texttt{TDB} with either multi-step objective or next encoding prediction (a) retains the nearly perfect test accuracy of vanilla sequence models (b) consistently solves the shortest paths problems when paired with an external solver---while both transformer or LSTM catastrophically fail---(c) learns cognitive maps nearly isomorphic to the ground truth.}
\label{table:aliased}
\end{table*}

\medskip

\textbf{Results: }
Table \ref{table:aliased} averages the metrics over $10$ experiments: each run considers a different GT room. First, we observe that both vanilla LSTM and transformer almost perfectly predict the next observation given the history. However, both sequence models struggle to solve path planning problems with forward rollouts: they only improve over the fallback path $\sim30\%$ of the same. When they do so, they find shorter paths $16$ times longer than an optimal path\footref{note1}. 

\begin{figure*}[!t]
\centering
    \begin{tabular}{c}
        \includegraphics[height=0.085\textheight]{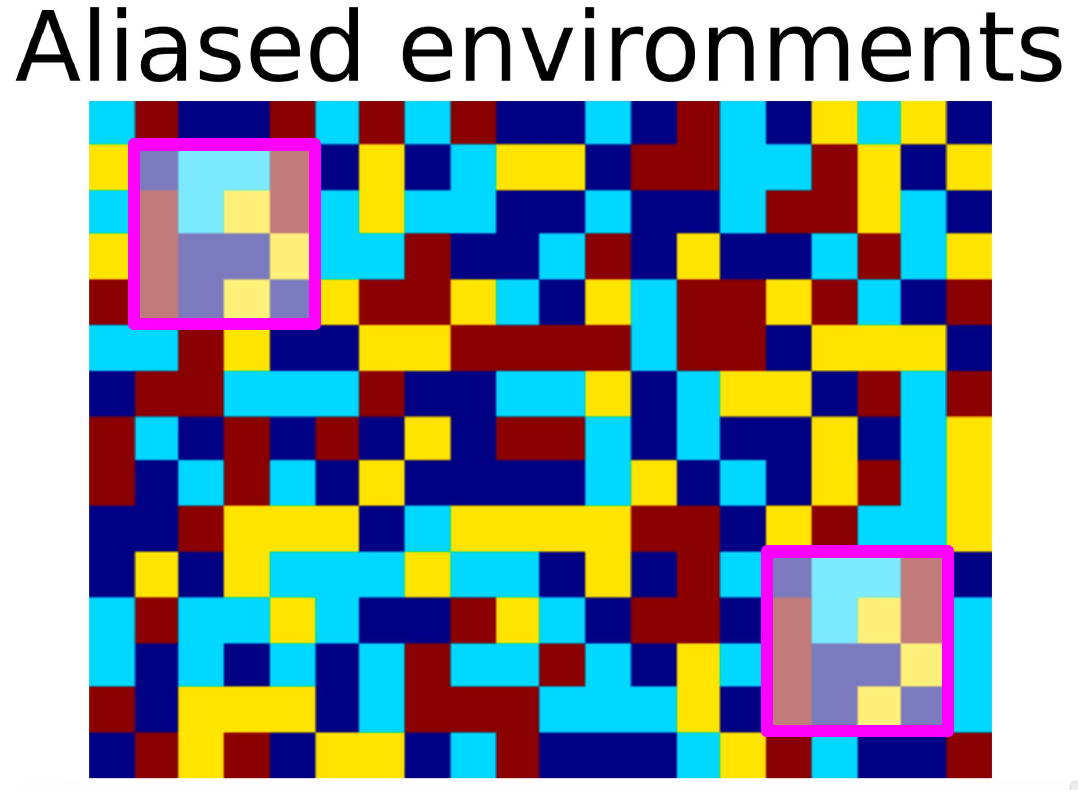}
        \includegraphics[height=0.085\textheight]{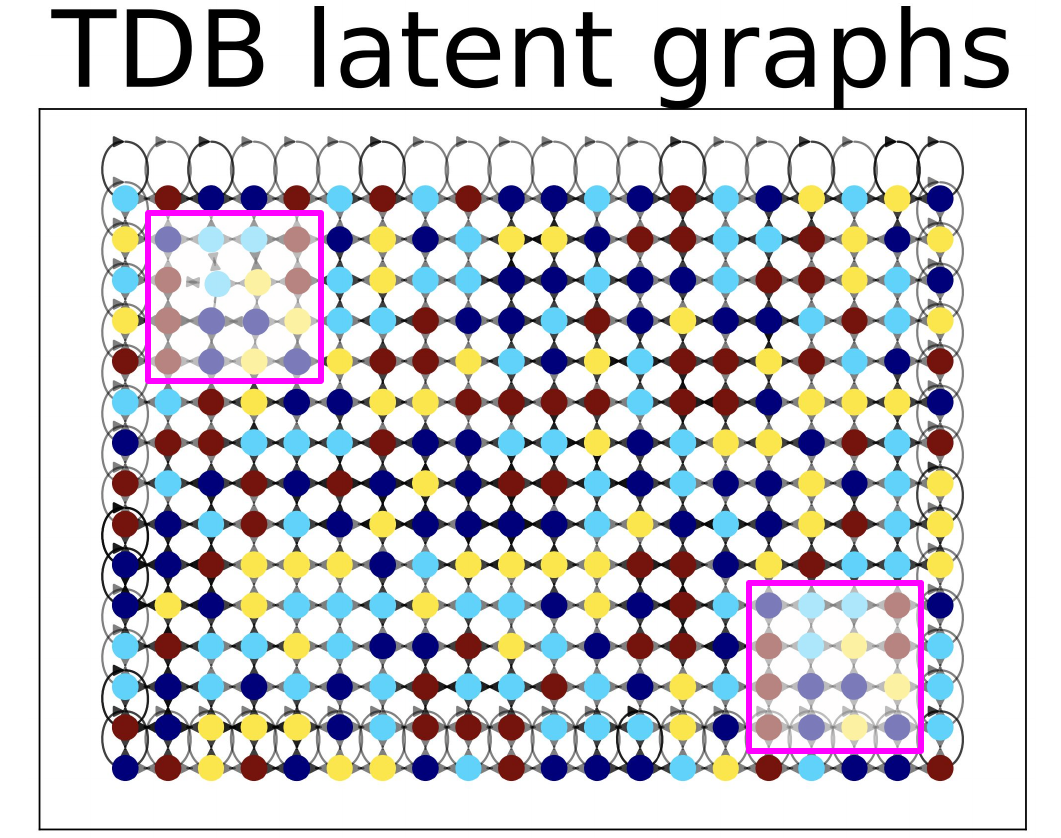}\\
        \includegraphics[height=0.095\textheight]{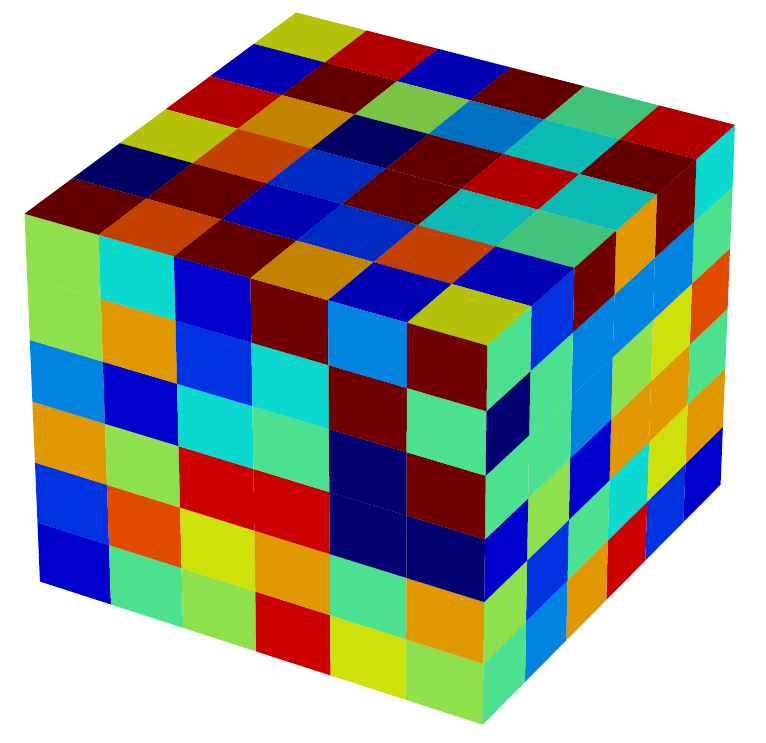}
        \hspace{1em}
        \includegraphics[height=0.095\textheight]{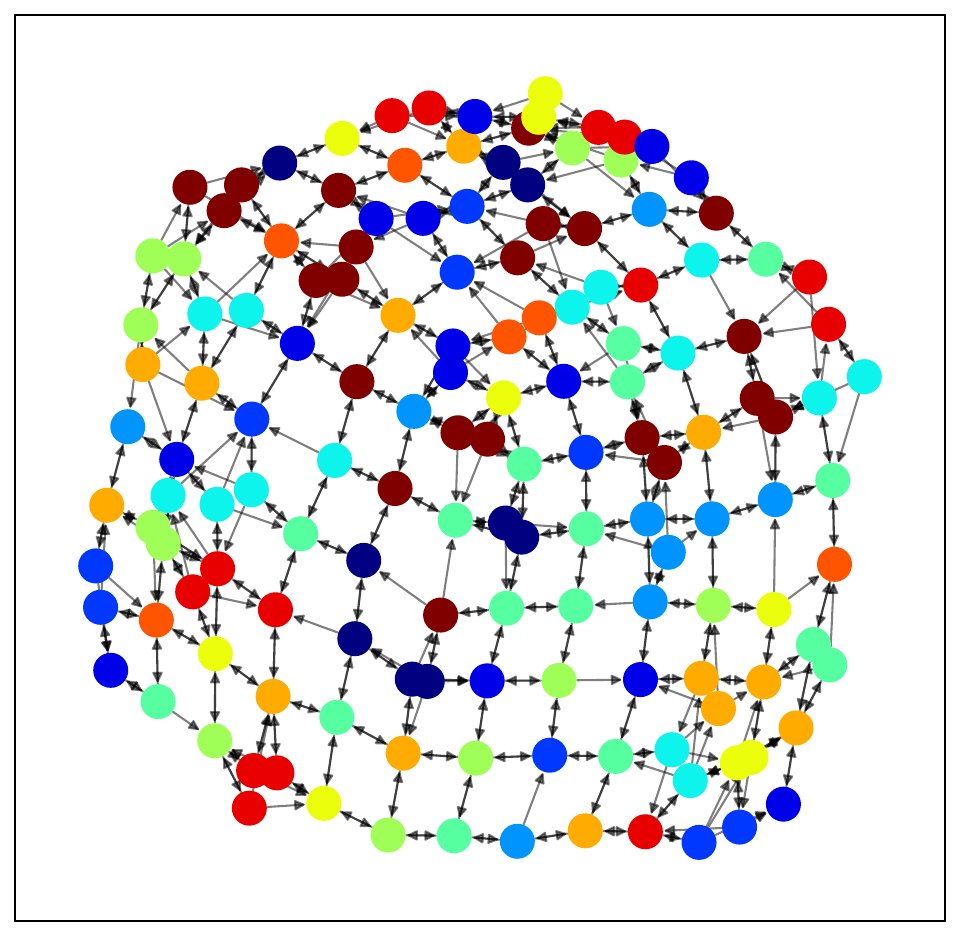}
    \end{tabular}
    \hspace{-1.5em}
    \begin{tabular}{c}
        \includegraphics[height=0.19\textheight]{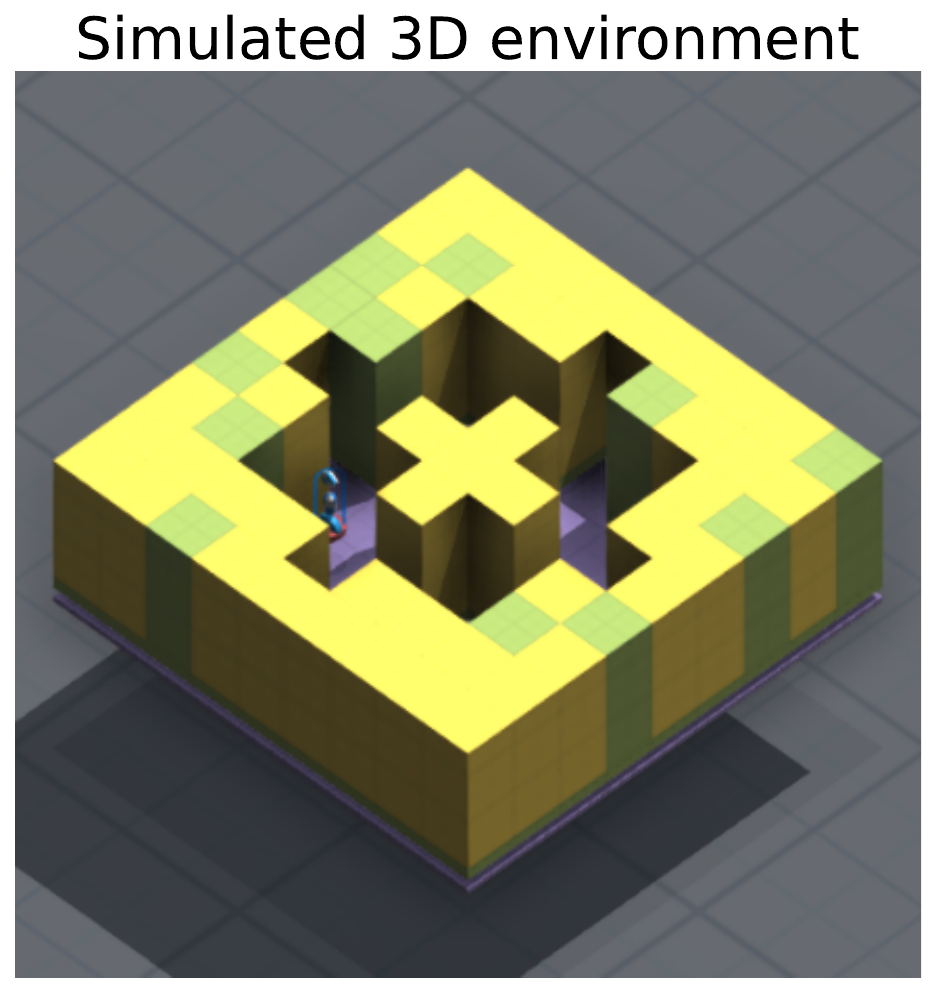}
    \end{tabular}
    \hspace{-1.8em}
    \begin{tabular}{c}
        \includegraphics[height=0.19\textheight]{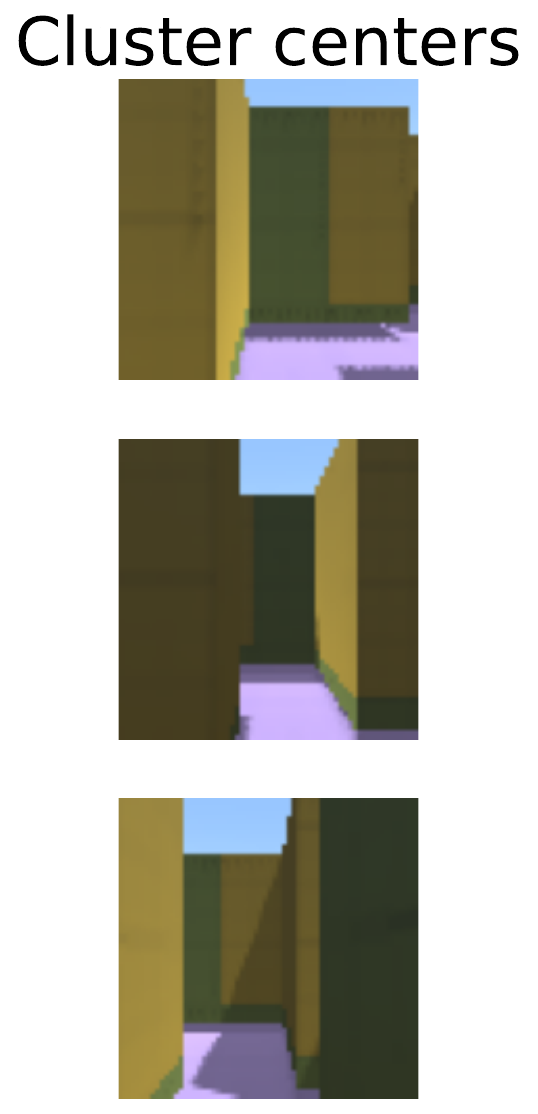}
    \end{tabular}
    \hspace{-1.7em}
    \begin{tabular}{c}
        \includegraphics[height=0.19\textheight]{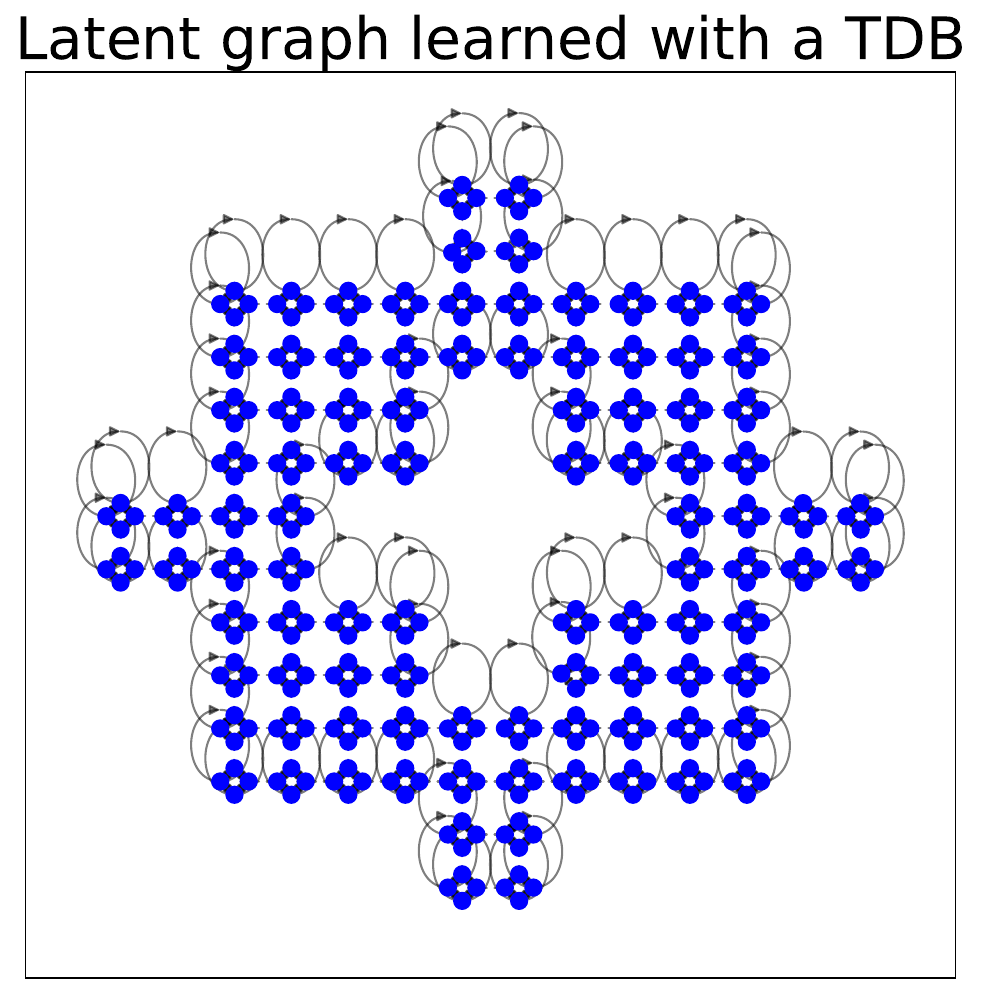}
    \end{tabular}
    \caption{[Left] Top: $2$D aliased room of size $\small{15\times20}$ with $O=4$ unique observations and $2$ identical $4\times4$ patches (in fuchsia). Bottom: Aliased cube of edge size $6$ with $O=12$ and non-Euclidean dynamics. [Center left] Top: cognitive map learned with a TDB($S=3, M=1$). For visualization, each latent node is mapped with the observation (resp. is placed at the $2$D GT spatial position) with higher empirical frequency when this node is active. Bottom: similar, but we use the Kamada-Kawai algorithm.
    [Center] Isometric view of a simulated $3$D environment. The agent navigates with egocentric actions and collects RGB images. [Center right] Three cluster centers: the cluster indices serve as categorical observations. [Right] Cognitive map learned with a TDB($S=3, M=4$): each location is represented by four nodes in the latent graph, corresponding to the four agent's heading directions. 
    Also see Fig.\ref{fig:agix_appendix}, Appendix \ref{appendix:agix_environment}. 
    }
    \label{fig:agix_results}
\end{figure*}

Second, all the \texttt{TDB}s retain the predictive performance of vanilla models. However, \texttt{TDB}$\small{(S=1, M=1)}$ fails at path planning: it only finds better valid paths $7\%$ of the time as it only solves the simplest problems. Indeed, the learned cognitive maps merge the observations with identical neighbors, which introduces unrealistic \textit{shortcuts} in the latent graphs---see Fig.\ref{fig:appendix_merging_single_step_aliased}, Appendix \ref{appendix:merging_single_step}. In contrast, \texttt{TDB} with either (a) multi-step or (b) next encoding objective learns cognitive maps that recover the GT dynamics and reach low \texttt{NormGED}---see Fig.\ref{fig:agix_results}[center left]. They are able to almost consistently find an optimal shortest path. As a result, \texttt{ImpFallback} is above $99\%$ while \texttt{RatioSP} is $1.00$. 

\medskip

\textbf{Multiple discrete bottlenecks accelerate training: } Table \ref{table:appendix_faster_trainig_aliased}, Appendix \ref{appendix:faster_training}, reports the average number of training steps to reach a $98\%$ train accuracy: for $\small{S=3}$, this number goes from $8300$ for $\small{M=1}$ down to $4000$ for $\small{M=4}$. In addition, multiple discrete bottlenecks seem more robust to the environment. Table \ref{table:aliased_appendix}, Appendix \ref{appendix:aliased_environment_12}, shows that, when the room contains $\small{O=12}$ unique observations, all the \texttt{TDB}s with $\small{M=4}$ outperform their $\small{M=1}$ counterparts. 

\medskip

\textbf{Lack of disentanglement in the latent space: }
We test whether multiple discrete bottlenecks learn a disentangled representation by training a logistic regression to predict each bottleneck index given the other indices---see Appendix \ref{appendix:disentanglement} for details. If the bottlenecks were independent, test disentanglement accuracy would be $0.20\%$. However, it is $84.99\%(\pm 0.56\%)$, which shows the strong redundancy in the representations learned by each bottleneck.

\medskip

\textbf{The agent can locate itself after a short context: }
Appendix \ref{appendix:aliased_environment_4} first shows that, on test random walks, accuracy is initially low then quickly increases: it is above $99.50\%$ after the $30$th observation---see Fig.\ref{fig:appendix_prediction_accuracy_by_obs}. 
Second, latent codes have a bimodal distribution: many codes with low empirical frequencies model the agent's early uncertainty and are replaced by high-frequency latent codes which later express the agent's confidence---see Fig.\ref{fig:appendix_counts_vs_obs}. 
Finally, the agent can ``teleportate'' in the room but such behavior appears less than $1\%$ of the time after the $60$th observation---see Fig.\ref{fig:appendix_teleportation}.

\medskip

\textbf{TDB can solve constrained path planning problems: }
\texttt{TDB} can inject constraints on demand into the external solver. To illustrate this, we consider a path planning problem variant: the new task is to find the shortest path from $\pos_C$ to $\pos_{N-C}$ which avoids, when possible, a randomly picked color. \texttt{TDB}$\small{(S=3, M=1)}$ is still able to almost consistently solve the problem: \texttt{ImpFallback} is still $99.55\% (\pm 0.13\%)$ while \texttt{RatioSP} is still $1.00 (\pm 0.00)$. The average length of the shortest paths is now $13.13 (\pm0.23)$, as opposed to $10.73(\pm0.08)$ when all the colors are allowed.

\medskip

\textbf{TDB handles noise during training: } When replacing $10\%$ of the training observations of a \texttt{TDB}$\small{(S=3, M=1)}$ by random ones, training accuracy drops to $88.03\% (\pm 0.04\%)$, while \texttt{TestAccu}, \texttt{ImpFallback} and \texttt{RatioSP} are still $\small{99.08\% (\pm 0.01\%), 99.49\% (\pm 0.20\%)}$ and $\small{1.00\% (\pm 0.00\%)}$, showing that training noise does not affect the test prediction and planning performance.

\medskip

\textbf{TDB learn maps for non-Euclidean dynamics: }
\texttt{TDB} can be applied to aliased environments with non-Euclidean dynamics. We generate data from a $3$D aliased cube with edge size $6$ and $O=12$ unique observations---see Fig.\ref{fig:agix_results}[bottom left]. Table \ref{table:aliased_cube}, Appendix \ref{appendix:aliased_cube} shows that the prediction and planning performance of a \texttt{TDB} are near perfect. Fig.\ref{fig:agix_results}[bottom - center left] visualizes the learned graph with the Kamada-Kawai algorithm \citep{kamada1989algorithm},

\subsection{Egocentric views in 3D simulated environments}\label{sec:exp_agix}

\textbf{Problem: } We consider a suite of visually rich $3$D simulated environments \citep{beattie2016deepmind} with a similar setting as \citet{guntupalli2023graph}---see Fig.\ref{fig:agix_results}[center, center right]. At each step, the agent can take any of three discrete \textit{egocentric} actions---move $1$m forward, rotate $90\deg$ left, rotate $90\deg$ right---and it sees a new view, which is a $64\times64$ RGB image. The agent's egocentric views are clustered with a $k$-means quantizer---with $\small{k=128}$--- and the clusters indices are used as categorical observations. The training and test set sizes are identical to Sec.\ref{sec:exp_aliased_room}.

\medskip

\textbf{Training: } We train the same models as in  Sec.\ref{sec:exp_aliased_room}, with the same parameters, on $10$ synthetic $3$D rooms.

\medskip

\textbf{Results: } The full results are reported in Table \ref{table:agix_room}, Appendix \ref{appendix:agix_environment}. Despite their excellent predictive performance, vanilla sequence models only find a path better than fallback for $26.60\% (\pm 0.60\%)$ of the problems: these ``better'' paths are $16.38 (\pm 0.87)$ times longer than optimal paths\footnote{\label{note1}The transformer and LSTM performance are almost identical because the same random walks are used for both models.}. Compared with  Sec.\ref{sec:exp_aliased_room}, training \texttt{TDB}s with a single bottleneck do not converge. Consequently, these models reach a low test accuracy and cannot solve the path planning problems. In contrast, for \texttt{TDB}s with $\small{M=4}$ discrete bottlenecks, averaged train accuracies are at least $99.70\%$ after $1000$ only training steps (see Table \ref{table:appendix_faster_trainig_agix}, Appendix \ref{appendix:faster_training}). These models reach nearly perfect (a) test accuracy and (b) path planning performance. Their latent maps accurately model the POE's dynamics---see Fig.\ref{fig:agix_results}[right] and Fig.\ref{fig:agix_appendix}, Appendix \ref{appendix:agix_environment}. However, the multiple bottlenecks learn a highly redundant representation: the test disentanglement accuracy (see Appendix \ref{appendix:disentanglement}) of a \texttt{TDB}$(S=3, M=4)$ is $99.06\%~(\pm0.06\%)$.

\subsection{Extension to text via the GINC dataset}

\textbf{Problem:} 
Herein, we extend \texttt{TDB} to text datasets to show that it can extract interpretable latent graphs on this other modality. We consider the text GINC dataset, introduced in \citet{xie2021explanation} to study in-context learning (ICL). GINC\footnote{available at \small{\url{https://github.com/p-lambda/incontext-learning/tree/main/data}}.} is generated from a uniform mixture of five factorial HMMs \cite{ghahramani1995factorial}---referred to as a \textit{concepts}. Each concept has two factorial chains with $10$ states each. Each state can emit $50$ observations shared across concepts. The training set consists of $1000$ training documents with a total of ${\sim}10$ million tokens: each document uniformly selects a concept and samples independent sentences from it. The test set consists of in-context prompts. Each prompt has between $n=0$ and $n=64$ examples: each example is of length $k \in \{3,5,8,10\}$. There are $2500$ prompts for each setting $(k,n)$. Each prompt uniformly selects a concept, samples $n-1$ examples $x^{(1)}_{:k}, \ldots, x^{(n-1)}_{:k}$ of length $k$, and one example $x^{(n)}_{:k-1}$ of length $k-1$. The in-context task is to infer the most likely last token of the last example, i.e., $\argmax_{x_{k-1}^{(n)}} p\left(x_{k-1}^{(n)} | x^{(1)}_{:k}, \ldots, x^{(n-1)}_{:k}, x^{(n)}_{:k-1}\right)$.

Since the vocabulary is shared among the concepts, observations in GINC are aliased like in natural language. Solving the task requires the model to disambiguate the aliased observations and correctly infer the latent concepts.

\medskip

\textbf{Training:} We train the same models as above using the same parameters, except the batch size which we set to $24$.

\medskip

\begin{figure*}[!t]
    \centering
    \includegraphics[width=0.69\textwidth]{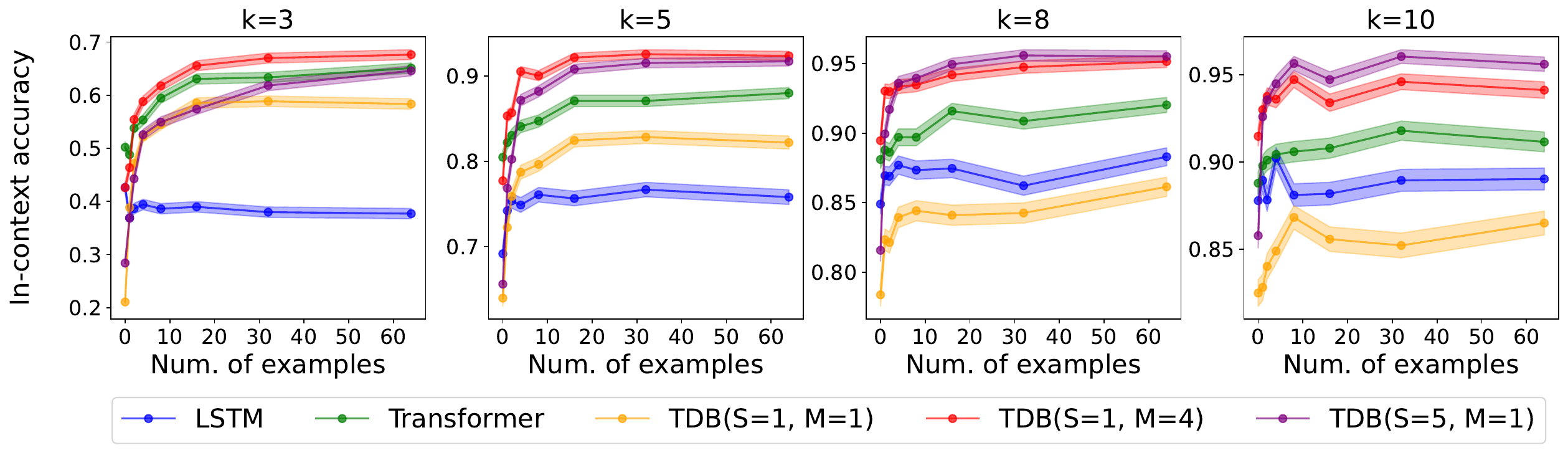}
    \hspace{0.02\textwidth}
    \includegraphics[width=0.27\textwidth]{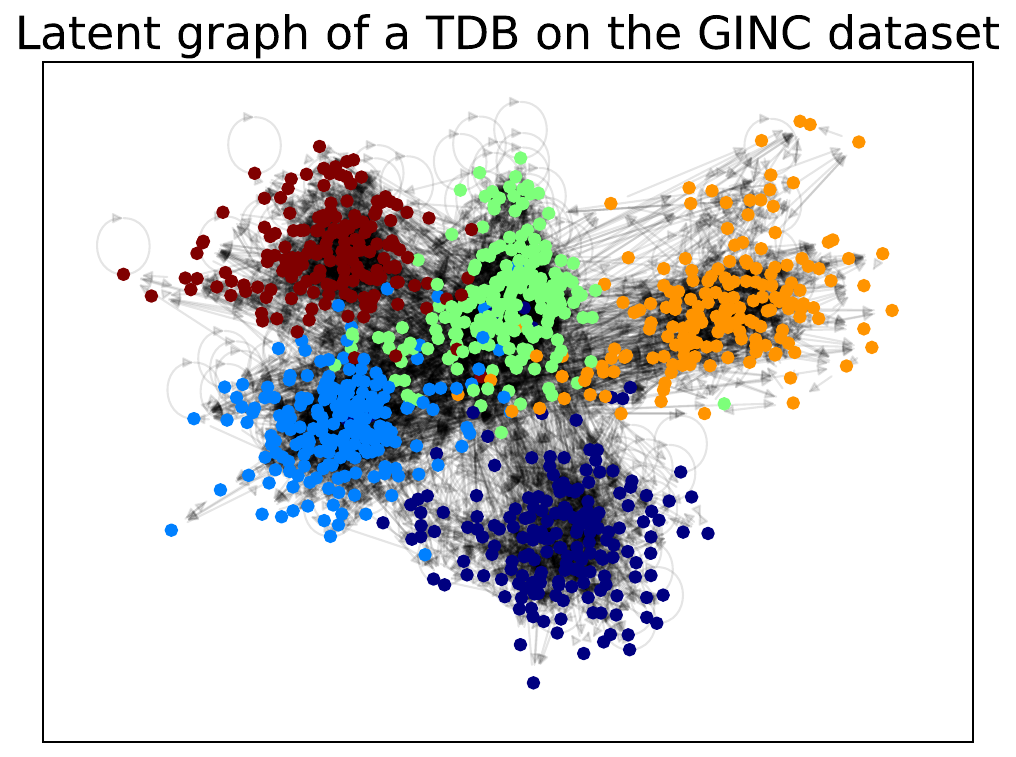}
    \caption{[Left] For all context lengths $k$, \texttt{TDB}($S=1, M=4$) achieves higher in-context accuracies than an LSTM and a vanilla transformer on the GINC test dataset \citep{xie2021explanation}---while \texttt{TDB}($S=5, M=1$) is the best model for large contexts. [Right] The learned latent graph of \texttt{TDB}($S=5, M=1$) exhibits five clusters, each corresponding to a color-coded concept.}
    \label{fig:ginc_results}
\end{figure*}

\textbf{Results: } Fig.~\ref{fig:ginc_results}[left] reports the in-context accuracy---defined as the average ratio of correct predictions---for each pair $(k,n)$ of the GINC test set. $\texttt{TDB}(S=1, M=1)$ is outperformed by vanilla sequence models. However, $\texttt{TDB}(S=1, M=4)$ outperforms both transformer and LSTM for all context lengths $k$. For larger contexts $k\in \{8,10\}$, the best model is $\texttt{TDB}(S=5, M=1)$---it is however slower to converge (see Table \ref{table:appendix_faster_trainig_ginc}, Appendix \ref{appendix:faster_training}). Finally, the highest in-context accuracies for \texttt{TDB} are around $95\%$ and are higher than what was reported in \citet{xie2021explanation} for transformers. The numerical values are in Table \ref{table:appendix_ginc}, Appendix \ref{appendix:ginc}.

Fig.~\ref{fig:ginc_results}[right] displays the latent graph learned by \texttt{TDB}$(S=5, M=1)$. We use $t_{\text{ratio}}=0.001$ and the Kamada-Kawai algorithm. Each latent node has a color corresponding to the GT concept with higher empirical frequency when this node is active. The learned latent graph has an interpretable structure: \texttt{TDB} learns five latent subgraphs corresponding to the five concepts in the GINC dataset. 

\begin{figure*}[!t]
    \centering
    \begin{tabular}{c}
        \includegraphics[height=0.148\textheight]{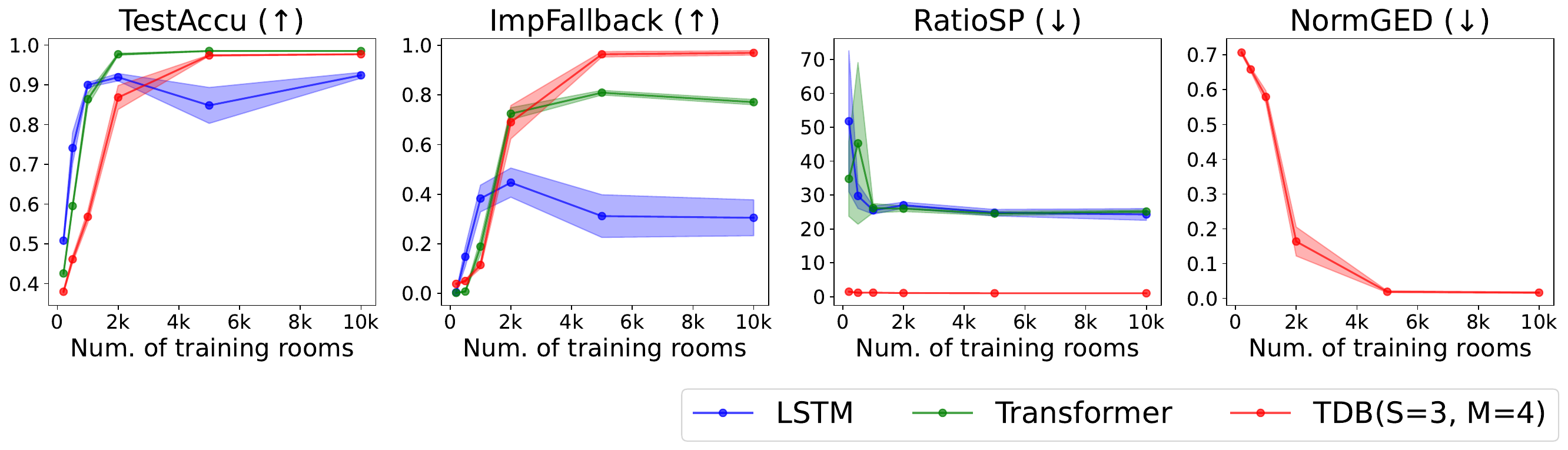}
    \end{tabular}
    \begin{tabular}{c}
        \includegraphics[height=0.077\textheight]{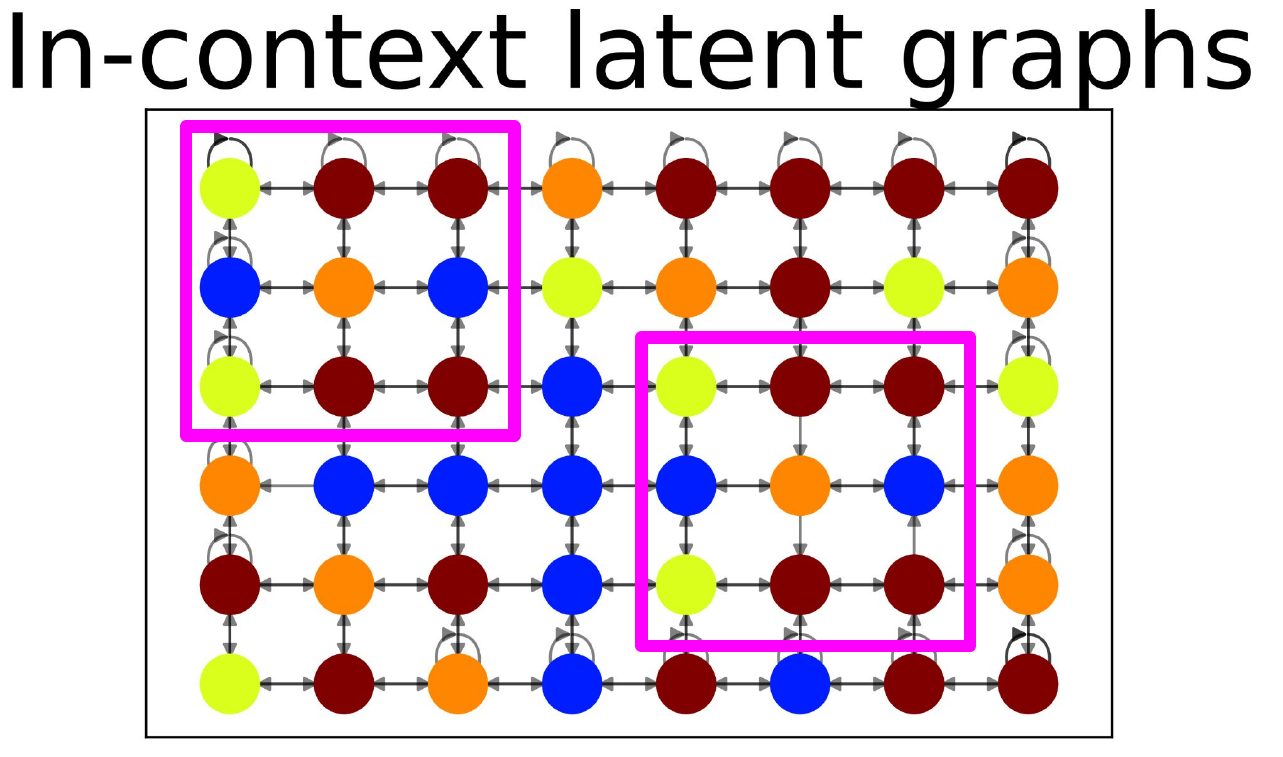}\\
        \includegraphics[height=0.065\textheight]{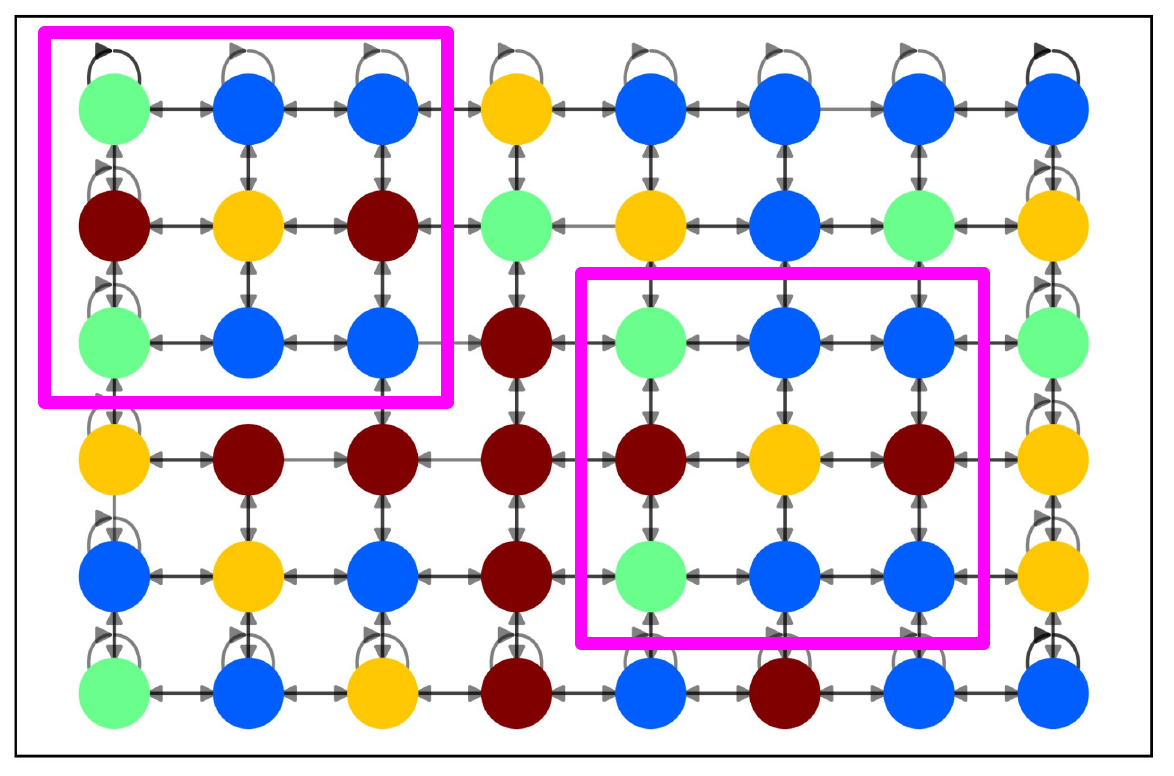}
    \end{tabular}
    \caption{[Left] In \textit{novel} $2$D aliased test rooms, \texttt{TDB}$(S=3, M=4)$ perfectly (a) predicts the next observation (b) solves in-context path planning problems. These in-context capacities emerge when the number of training rooms increases. A vanilla transformer only solves the prediction problem (a)---which an LSTM struggles to do. [Right] Two latent graphs in-context learned by the \texttt{TDB} on two new test rooms. By design (a) a $3\times3$ fuchsia-coded patch is repeated twice and (b) the room partitions induced by the colors are the same.}
    \label{fig:icl_results}
\end{figure*}

\subsection{In-context learning experiments}\label{sec:exp_icl}

\textbf{Problem: } Our last experiment explores whether vanilla models and \texttt{TDB} can, in a new test POE, (a) in-context predict the next observation (b) solve in-context path planning problems (c) learn in-context latent graphs. Similar to Sec.\ref{sec:exp_aliased_room}, we first generate a $2$D aliased room of size $6 \times 8$ with $O=4$ unique colors, and with global aliasing: a $3\times3$ patch is repeated twice. We define the \textit{room partition} as the partition $\mathcal{S}_1, \ldots, \mathcal{S}_O$ of the GT room spatial positions such that all the room spatial positions in a set $\mathcal{S}_i$ have the same observation. We then generate many aliased rooms of the same size by preserving the room partition: each room picks $4$ colors out of $30$ without replacement, and assigns all the positions in $\mathcal{S}_i$ to the same color. See Appendix \ref{appendix:icl_setup}.

We train each model for a varying number of training rooms: $200, 500, 1\text{k}, 2\text{k}, 5\text{k}, 10\text{k}$ \footnote{That is, we train on at most $1.52\%$ of all the possible rooms.}. The training set contains $8092$ sequences: each sequence picks a training room at random, then generates a random walk of length $400$ in it. The test set only contains four random walks in each \textit{new} test room. We define in-context path planning problems as before, using a context $C=100$. Given a new test room, \texttt{TDB} derives the test bottleneck indices, builds an in-context cognitive map, and uses it to solve the in-context path planning problems.

\medskip

\textbf{Training in base: } We map a sequence of observations $x=(x_1, x_2,\ldots,x_N)$ in a room---where $x_n$ is one of the four room observations---to a room-agnostic sequence $\tilde{x}=(\tilde{x}_1,\ldots,\tilde{x}_N)$ where (a) $\tilde{x}_n \in \{1, \ldots, O\}$ (b) $\tilde{x}_n=k$ iff. $\tilde{x}_n$ is the $k$th lowest observation seen between indices $1$ and $n$\footnote{$\tilde{x}_n=1$ may map to different $x_n$ through the sequence.}. We train each model to take $x$ as input and to predict $\tilde{x}$. Targets are always in the \textit{base} $\{1, \ldots, O\}$, which encourages the models to share structure across rooms---e.g. \texttt{TDB} can reuse the same latent codes across rooms. As before, we train for $25$k Adam iterations, using a learning rate of $0.001$, a batch size of $32$ and a dropout of $0.1$.

\medskip

\textbf{Results: } Fig.~\ref{fig:icl_results}[left] averages the results over $10$ runs---the numerical values are in Table \ref{table:appendix_icl}, Appendix \ref{appendix:icl_results}. For \texttt{NormGED}, we use a timeout of $20s$. As the number of training rooms increases, all the models display a phase transition where both prediction and path planning performance improve. When the number of training rooms is larger than $5000$, in-context learning \textit{emerges} for the transformer and \texttt{TDB}: both achieve almost perfect in-context accuracy on the \textit{new} test rooms. In contrast, LSTM reaches lower in-context accuracy.
In addition, in-context path planning emerges in the \texttt{TDB}: it can nearly perfectly solve the shortest paths problems, which both vanilla models struggle with. The in-context latent graphs reach low \texttt{NormGED} and are nearly isomorphic to the GT, and correctly model the room's dynamics---see Fig.~\ref{fig:icl_results}[right]. \texttt{TDB}$(S=3,M=4)$ trained on $10$k rooms has a disentanglement accuracy (see Appendix \ref{appendix:disentanglement}) of $94.02\% (\pm0.27\%)$: again, multiple discrete bottlenecks do not learn a disentangled representation. Finally, Appendix \ref{appendix:icl_spatial} studies how spatial exposure to base targets in the training data drives in-context performance.


\section{Conclusion}

We propose \texttt{TDB}, a transformer variant that addresses the shortcomings of vanilla transformers for path planning. \texttt{TDB} (a) introduces discrete bottlenecks which compress the information necessary to predict the next observation given history then (b) builds interpretable cognitive maps from the active bottlenecks indices. On perceptually aliased POEs, \texttt{TDB} (a) retains the near-perfect prediction of transformers, (b) calls an external solver on its latent graph to solve path planning problems exponentially faster, (c) learns interpretable structure from text datasets, (d) exhibits emergent in-context prediction and path planning abilities.

Our approach has two main limitations. First, \texttt{TDB} only accepts categorical inputs. Second, though multiple discrete bottlenecks accelerate training, they do not learn a disentangled latent space. 
To address these points, we want to modify the \texttt{TDB} architecture (a) to accept high-dimensional continuous observations (images) (b) so that different bottlenecks learn non-redundant representations. 
Furthermore, we would like to extend this work into a framework to build planning-compatible world models in rich environments. To do so, we want to learn disentangled latent dynamics in factored Markov decision processes by (a) extracting knowledge with transformers, (b) using multiple discrete bottlenecks to compress this knowledge and to generate latent nodes---or local latent graphs---and (c) learning factorized transition matrices over these latent nodes.




\section*{Acknowledgments}
We thank Kevin Murphy, Daan Wierstra and Th\'eophane Weber for useful discussions during the preparation of this manuscript.

\newpage
\bibliographystyle{plainnat}
\bibliography{arxiv}

\newpage
\appendix
\onecolumn

\section{Vanilla transformer architecture} \label{sec:appendix_vanilla_transformer}

\begin{figure*}[!ht]
    \centering
    \includegraphics[width=.8\textwidth]{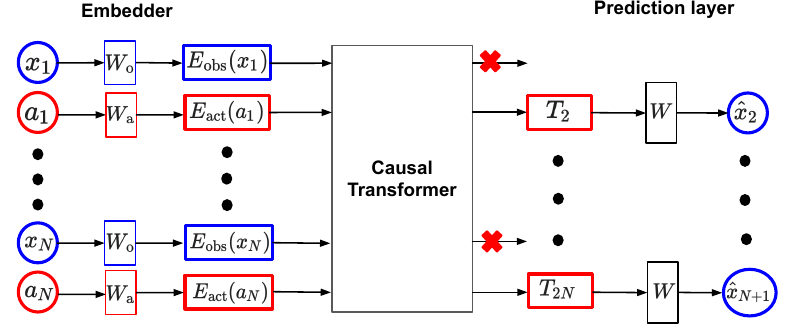}
    \caption{A vanilla transformer with causal mask takes as inputs the respective linear embeddings of the observations and actions. A linear layer on top of the action representations returns the next observation logits.}
    \label{fig:vanilla_transformer}
\end{figure*}

Fig.\ref{fig:vanilla_transformer} presents the vanilla transformer detailed in Sec.\ref{sec:vanilla_transformer}. In our numerical experiments, Sec.\ref{sec:exp}, we train this model to minimize the autoregressive objective Equation \eqref{eq:transformer_objval}.

\newpage
\section{On the connection between TDB and CSCG}

\subsection{Computational and space complexity}\label{sec:appendix_cscg_complexity}

A clone-structured causal graph (CSCG) \citep{george2021clone} is a hidden Markov model variant used to model sequences of categorical observations. The latent space of a CSCG is partitioned, such that all the latent variables in one set of the partition deterministically emit the same observation. These latent variables are the \textit{clones} of the observation. Because of its latent structure, CSCG models a transition between two observations as a transition between their respective clones---as opposed to a transition between all the latent variables.

We compare below the complexity between a vanilla CSCG and a vanilla self-attention layer when running inference on a sequence of length $N$.

\medskip

\textbf{Computation complexity: } For a vanilla CSCG with $M$ clones per observation, the computational cost for computing the forward and backward messages is $O(M^2 N)$. In contrast, for a vanilla self-attention layer with a $D$-dimensional embedding, the forward pass cost is $O(N^2 D)$. In this paper, we use $N=400$, $D=256$. CSCG is then more expensive when the number of clones per observation satisfies $M \gg \sqrt{N D} = 320$. 

\medskip

\textbf{Space complexity: } The memory cost of storing a vanilla CSCG with $M$ clones per observation and $E$ different observations is $O(M^2 E^2)$. This large dense memory matrix can grow quickly as the number of observations increases, and prohibit vanilla CSCGs from scaling to very large latent space. In contrast, a vanilla self-attention layer with a $D$-dimensional embedding requires a $O(N^2 + N D)$ memory. The quadratic dependence on the sequence length is problematic for large corpora, but not for the problem we study here, where $N=400$.

\bigskip

\subsection{A transformer with clone-structured discrete bottleneck}\label{sec:appendix_clone_structure}

We propose herein a variant of the \texttt{TDB} architecture inspired by the CSCG model, which injects a ``clone-structured'' inductive bias into its discrete bottlenecks. We consider the dictionary of latent codes introduced in  Sec.\ref{sec:discrete_bottleneck}: $\mathcal{D} = (d_1, \ldots, d_K)$, where $d_k \in \mathbb{R}^D, ~\forall k$. We now allocate a subset of the latent codes for each observation $x$ in the vocabulary $V$. That is, we first define a partition of the indices $\{ \mathcal{C}(x) \}_{x \in V}$ such that: 
$$\bigcup_{x \in V} \mathcal{C}(x) = \{1, \ldots, K\} ~~~;~~~
\mathcal{C}(x) \cap \mathcal{C}(y) = \emptyset, ~\forall x \ne y ~~~;~~~
\mathcal{C}(x) \ne \emptyset, ~\forall x.$$ 
Second, we generalize the operator $\phi$ in Equation \eqref{eq:vector_quantize} so that it now depends on both (a) the transformer output and (b) the observation itself. That is, we define:
\begin{equation}
\tilde{\phi}(y, x) = \argmin_{k \in \mathcal{C}(x)} \| y - d_k \|_2^2, ~~ \forall y \in \mathbb{R}^d, ~ \forall x \in V
\end{equation}
For the observation $x_n$ which leads to the transformer output $T_{2n-1}$, the discrete quantizer now returns the latent code of $\mathcal{D}$ with index $\tilde{\phi}(T_{2n-1}, x_n)$. That is, each observation can now only activate a subset of the latent codes. We refer to this new quantizer module as \textit{clone-structured discrete bottleneck}.

\medskip

\textbf{Computational gain: } This architecture change decreases the computational cost of the discrete bottleneck by a linear term $O(E)$---which can be of interest when the number of observations $E$ is large. However, the overall computational cost is still dominated by the quadratic cost of computing the attention matrices in the causal transformer.

\medskip

\textbf{Performance gain: } Overall, we did not find any major benefit from replacing the discrete bottleneck of a \texttt{TDB} with its clone-structure variants. Our most promising result was that we were able to solve the $3$D synthetic environments in Sec.\ref{sec:exp_agix} using a single clone-structured discrete bottleneck---where a \texttt{TDB} needs more than one discrete bottleneck. However, a transformer with clone-structured bottlenecks would struggle to solve the in-context learning experiments in Sec.\ref{sec:exp_icl}. Indeed, a new test room with observations $5, 6, 7, 8$ would, by definition, use different latent codes than a training room with observations $1, 2, 3, 4$, which prevents the model from learning to share structure across rooms. This convinced us to not use the clone-structured discrete bottleneck in our experiments.

\newpage
\section{Clustering steps to build the transition graph} \label{sec:appendix_cluster}
In this section, we detail the clustering steps that we apply to learn a cognitive map from the bottleneck indices in Sec.\ref{sec:world_model}.

\paragraph{Step 1. Cluster the bottlenecks indices with Hamming distance: }
This first clustering step only applies in the case where the \texttt{TDB} has multiple $M>1$ discrete bottlenecks. In this case, \texttt{TDB} maps each observation $x_n$ to a tuple of latent code indices:
$$
s_n=\left(\phi^{(1)}(T_{2n-1}), \ldots, \phi^{(M)}(T_{2n-1}) \right) \in \{1, \ldots, K\}^M.
$$
Before building the cognitive map, we first collect in a set $\mathcal{J}$ all the unique tuples of latent indices appearing in the training set. For two tuples $s=(s^{(1)},\ldots,s^{(M)}) \in \mathcal{J}$ and $\ell=(\ell^{(1)},\ldots,\ell^{(M)}) \in \mathcal{J}$, we compute their Hamming distance:
$$
H(s, \ell) = \frac{1}{M} \sum_{j=1}^M \mathbf{1}(s^{(j)} \ne \ell^{(j)}) \in [0, 1].
$$
We introduce a distance threshold $d_{\text{Hamming}}$---which we set to $0.25$ in Sec.\ref{sec:exp}---and merge any pair of tuples such that $H(s, \ell) \le d_{\text{Hamming}}$. This means that we will not distinguish $\ell$ and $s$ when we build the count tensor $C$---see Sec.\ref{sec:world_model}. Consequently, these two tuples will be mapped to the same node on the cognitive map $\mathcal{G}$---and any node in $\mathcal{G}$ will be associated with a \textit{collection of tuples} of bottleneck indices. 

Intuitively, this means that when the entries of two tuples of bottleneck indices are ``almost all the same'', then we assume that their associated observations are at the same spatial position.

\paragraph{Step 2. Merge identical nodes: } Second, we build the count tensor $C$ (after this first clustering step), and threshold it with $t_{\text{ratio}}$ to build an action-augmented transition graph $\mathcal{G}$---as detailed in Sec. \ref{sec:world_model}. We only retain nodes that have at least two incoming and outgoing neighbors. Finally, our second clustering step merges every pair of nodes in $\mathcal{G}$ that are connected to the same set of neighbors.

\paragraph{Step 3. Map each discarded node to its closest retained node: } Thresholding the count tensor $\mathcal{C}$ may discard some (tuple of) bottleneck indices with low empirical counts. In particular, as we show in Fig.\ref{fig:appendix_counts_vs_obs}, Appendix \ref{appendix:aliased_environment_4}, many bottleneck indices that are active early in the sequences are discarded. This may be problematic for solving path planning problems, as it means that the corresponding observations cannot be mapped to nodes in $\mathcal{G}$. Indeed, when \texttt{TDB} activates a discarded (tuple of) bottleneck indices, it cannot locate itself in the learned cognitive map $\mathcal{G}$.

To remedy this, we divide the set of unique (tuples of) latent indices $\mathcal{J}$, between the elements that are discarded---i.e. not mapped to a node in $\mathcal{C}$---and retained---i.e. mapped to a node in $\mathcal{C}$. Our last clustering step maps any discarded (tuple of) latent indices $s_{\text{discarded}}$ to its closest retained latent index for the $\ell_1$ distance, defined as:
$$
\hat{s}_{\text{retained}} ~~\in~~ \argmin_{s_{\text{retained}}}
\sum_{a=1}^{N_{\text{actions}}} 
\sum_{\ell=1}^K
\left| p( \ell ~|~ s_{\text{retained}}, a) - p( \ell ~|~ s_{\text{discarded}}, a) 
\right|.
$$
After this step, $s_{\text{discarded}}$ and $\hat{s}_{\text{retained}}$ are mapped to the same node in $\mathcal{C}$, and all the entries in $\mathcal{J}$ are retained.

\newpage
\section{Solving path planning problem at test time on partially observed environments}\label{appendix:planning_metric}

We detail herein how a vanilla sequence model and a \texttt{TDB} can solve the test shortest path problem presented in Sec.\ref{sec:exp_aliased_room}.

\paragraph{Defining the shortest path problem: } We consider a test sequence of observations $x=(x_1, \ldots, x_N)$ and actions $a=(a_1, \ldots, a_N)$ in a POE---the environment can either be the $2$D aliased rooms of Sec.\ref{sec:exp_aliased_room} and Sec.\ref{sec:exp_icl}, the aliased cube of Sec.\ref{sec:exp_aliased_room}, or the egocentric $3$D synthetic environments of Sec.\ref{sec:exp_agix}. We also denote $\pos = (\pos_1, \ldots, \pos_N)$ the \textit{unobserved} sequence of room spatial positions associated with each observation. Note that, because of aliasing, we cannot deterministically infer $\pos_n$ from $x_n$.

For the $2$D aliased rooms (Sec.\ref{sec:exp_aliased_room}), the spatial position of an observation is its $2$D coordinates. For the egocentric $3$D synthetic environments (Sec.\ref{sec:exp_agix}), the spatial position of an observation is a $3$D vector: the first two entries correspond to the $2$D coordinates of the observation, while the last entry corresponds to the agent's heading direction, which is one of $\{0^{\circ}, 90^{\circ}, 180^{\circ}, 270^{\circ} \}$

Let $C$ be a context length---we set $C=50$ or $C=100$ in practice. The path planning problem we consider in Sec.\ref{sec:exp} consists of finding the shortest path, i.e., the shortest sequence of actions, which leads from the room position $\pos_C$ (associated with the observation $x_C$) to the room position $\pos_{N-C}$ (associated with the observation $x_{N-C}$). Naturally, the sequence $(a_C,\ldots, a_{N-C-1})$ is a path of length $N-2C$ from $\pos_C$ to $\pos_{N-C}$, which we refer to as the \textit{fallback path}. In addition, for evaluating a model's performance, we use an \textit{optimal shortest path}, which solves the shortest path problem in the ground truth room---note that the agent does not know the ground truth layout. While the optimal shortest path is not unique, all the optimal shortest paths have the same length.

Fig.\ref{fig:motivation} in the main text illustrates this path planning problem on a $2$D aliased room---we use $N=40, C=5$ for the sake of visualization.


We detail below our proposed procedure for solving this shortest path problem with a \texttt{TDB}; and with a vanilla transformer or a vanilla LSTM. As mentioned in the main text, a vanilla sequence model finds this challenging because (a) it can only perform forward rollouts---without being able to collapse redundant visitations of the same spatial position---and (b) it cannot tell whether it has reached its destination. In contrast, a \texttt{TDB} solves this problem by querying its learned latent graph.

We are interested in measuring (a) whether each model can derive a path that improves over the fallback path and (b) when a model derives such a better path, how much worse it is compared to the optimal shortest path.

\paragraph{Deriving shortest paths with TDB: } When \texttt{TDB} is evaluated on the test sequences $x$ and $a$, it maps each observation $x_n$ to either (a) a single latent code index $s_n=\phi(T_{2n-1})\in \{1, \ldots, K\}$ when we have a single discrete bottleneck or (b) a tuple of latent codes indices $s_n=\left( \phi^{(1)}(T_{2n-1}), \ldots, \phi^{(M)}(T_{2n-1}) \right)$ when we have multiple discrete bottlenecks.

As a reminder, each node in the learned cognitive map $\mathcal{G}$ is associated with a collection of (tuples of) bottleneck indices---see Appendix \ref{sec:appendix_cluster}. However, the (tuples of) bottleneck indices $s_C$ and $s_{N-C}$ may not be associated with a node in $\mathcal{G}$. Consequently, we find two tuples of bottleneck indices $\tilde{s}_C$ and $\tilde{s}_{N-C}$ (a) with lowest Hamming distance from $s_C$ and $s_{N-C}$ and (b) which are associated with two nodes $n_C$ and $n_{N-C}$ in $\mathcal{G}$ . Finally, we call the external \texttt{\small{networkx.shortest\_path}} solver to find a shortest path between the nodes $n_C$ and $n_{N-C}$.

\paragraph{Deriving shortest paths from rollouts with vanilla sequence models: } A naive option that finds the shortest path between $\pos_C$ and $\pos_{N-C}$ with a vanilla sequence model consists in exploring all the possible sequences of observations and actions. For a number of evaluations exponential in the length of the optimal shortest path, this approach is guaranteed to find an optimal shortest path.

We propose herein an alternative approach, which (a) is more computationally efficient, (b) is not guaranteed to find the optimal shortest path, and (c) in practice, often improves over the fallback path. Beforehand, let us define the subsequences of observations context $x_{\text{context}}=(x_1, \ldots, x_C)$ and actions context $a_{\text{context}}=(a_1, \ldots, a_{C-1})$. We also define the subsequences of observations tail $x_{\text{tail}}=(x_{N-C + 1}, \ldots, x_N)$ and actions tail $a_{\text{tail}}=(a_{N-C}, \ldots, a_{N-1})$.

First, we generate $N_{\text{random walks}}$ random sequences of actions, each of length $N-2C$: we denote $a^{(i)}=(a^{(i)}_C,\ldots, a^{(i)}_{N-C-1})$ the $i$th random sequence---we initialize the indices at $C$. In Sec.\ref{sec:exp_agix}, we set $N_{\text{random walks}} = 10$.

Second, we augment each sequence of actions (of length $N-2C$) into two sequences of observations and actions, of respective lengths $N-2C$ and $N-2C-1$, as follows. The first $C$ observations and $C-1$ actions are the shared context $x_{\text{context}}$ and $a_{\text{context}}$. The next $N-2C$ actions correspond to $a^{(i)}$. The corresponding sequence of observations corresponds to the sequence model's autoregressive predictions. That is, for $n = C,\ldots, N-C-1$, we iteratively define $\hat{x}^{(i)}_{n + 1}$ as:
$$
\hat{x}^{(i)}_{n + 1} \in \argmax_{x} p \left( x ~~\bigg|~~
\underbrace{x_1, ~a_1, \ldots, a_{C-1}, ~x_C}_{\text{shared context}}, ~a^{(i)}_{C}, ~\hat{x}^{(i)}_{C+1}, ~a^{(i)}_{C+1}, \ldots ,\hat{x}^{(i)}_{n}, ~a_{n}
\right)
$$
For each generated random walk, we look for all the observations equal to $x_{N-C}$, and define the set of \textit{candidates}:
$$\mathcal{C} = \left\{(i,n): n\ge C \text{ and } \hat{x}^{(i)}_n = x_{N-C} \right\}.$$

\paragraph{Estimating whether the target is reached via tail evaluation: }
As we mentioned in the main text, because of aliasing, the vanilla sequence model cannot know whether it has reached the target position when it observes a candidate. That is, it does not know whether the spatial position associated with the candidate is equal to $\pos_{N-C}$.

We could return all the candidates and evaluate them using the ground truth map. Instead, we propose herein to estimate whether a candidate $(i,n)$ is at the target position $\pos_{N-C}$ by evaluating it---with its history---on the tail subsequences of observations $x_{\text{tail}}$ and actions $a_{\text{tail}}$. That is, for a candidate $(i, n)$ we define, for $k = 1,\ldots,C$:
$$
\hat{x}^{(i)}_{\text{tail}, k} \in \argmax_{x} p \left( x ~~\bigg|~~
\underbrace{x_1, ~a_1, \ldots, a_{C-1}, ~x_C}_{\text{shared context}}, ~\underbrace{a^{(i)}_{C}, ~\hat{x}^{(i)}_{C+1}, ~a^{(i)}_{C+1}, \ldots ,\hat{x}^{(i)}_{n}=x_{N-C}}_{\text{autoregressive random walk}}, ~a_{N-C}, x_{N-C + 1}, \ldots, a_{N-C+k-1}
\right)
$$
We estimate that a candidate is at the target position $\pos_{N-C}$ when $(\hat{x}^{(i)}_{\text{tail}, 1}, \ldots, \hat{x}^{(i)}_{\text{tail}, C}) = x_{\text{tail}}$. Note that, because of aliasing, even when the tail evaluation succeeds, the candidate may not be at the target position.

\paragraph{Evaluating valid paths: }
For the \texttt{TDB}, we return a single shortest path proposal, which is estimated by the external solver. For the transformer and the LSTM, we return all the path proposals that are estimated to be at the target position. 

For each model, we first evaluate whether the proposed paths are \textit{valid}. That is, we test whether the proposed sequence of actions correctly leads from $\pos_C$ to $\pos_{N-C}$ in the ground truth environment. For a vanilla sequence model, when multiple returned paths are valid, we only retain the shortest one. We then compute two metrics \textbf{(a)} a binary indicator of whether the method has found a valid path and \textbf{(b)} if (a) is correct, the ratio between the length of the valid found path and the optimal path length. Finally, \texttt{ImpFallback} averages (a) while \texttt{RatioSP} averages (b) over $200$ shortest paths problems.

\paragraph{Tail evaluation improves the quality of the paths returned: } For a transformer trained on $2$D aliased rooms in Sec.\ref{sec:exp_aliased_room}, if we return all the candidates, then the ratio of \textit{valid} paths among the paths returned is $0.78\%(\pm 0.03\%)$. Indeed, most of the candidates are not at the target spatial position. When we estimate whether we are at the target position via tail evaluation, this ratio goes up to $38.76\%(\pm 1.17\%)$---prediction errors prevent it from being at $100\%$. Similarly, on the $3$D synthetic environments of Sec.\ref{sec:exp_agix}, a transformer without tail evaluation returns $1.75\% (\pm0.20\%)$ of valid paths while a transformer with tail evaluation returns $25.78\% (\pm 1.17\%)$. Finally, on the ICL experiments of Sec.\ref{sec:exp_icl}, for a transformer trained on $10k$ rooms, this same ratio goes from $2.58\% (\pm 0.05\%)$ without tail evaluation up to $51.10\% (\pm 1.81\%)$ with it. Overall, tail evaluation drastically improves the quality of the paths returned by sequence models.

\newpage
\section{Multiple discrete bottlenecks accelerate training}\label{appendix:faster_training}

Tables \ref{table:appendix_faster_trainig_aliased}, \ref{table:appendix_faster_trainig_agix} and \ref{table:appendix_faster_trainig_ginc} below illustrate that multiple discrete bottlenecks accelerate training.

\textbf{2D aliased rooms: } For each \texttt{TDB} trained on the $2$D aliased rooms and reported in Sec.\ref{sec:exp_aliased_room}, we compute, for every $1000$ training steps, its training accuracy on the entire training set. We define a new metric, \texttt{TrainingAbove98}, which reports the first evaluation when this training accuracy is higher than $98\%$. Table \ref{table:appendix_faster_trainig_aliased} averages this metric over the $10$ experiments run in Sec.\ref{sec:exp_aliased_room}, for various \texttt{TDB} models.

\begin{table*}[!ht]
\centering
\resizebox{0.45\textwidth}{!}{
\begin{tabular}{p{0.25\textwidth}p{0.25\textwidth} }
    \toprule
    Method & TrainingAbove98 $\downarrow$ \\
    \toprule
    \toprule
    \texttt{TDB}$(S=1, M=1)$ & $9400~(290)$ \\
    \texttt{TDB}$(S=1, M=4)$ & $4300~(145)$\\
    \midrule
    \texttt{TDB}$(S=1, \enc, M=1)$ & $9300~(425)$ \\
    \texttt{TDB}$(S=1, \enc, M=4)$ & $4500~(158)$\\
    \midrule
    \texttt{TDB}$(S=3, M=1)$ & $8300~(284)$\\
    \texttt{TDB}$(S=3, M=4)$ & $4000~(0)$\\
    \bottomrule
\end{tabular}}
\caption{On $2$D aliased rooms, training a \texttt{TDB} with $M=4$ discrete bottlenecks converges faster than training its counterpart with a single discrete bottleneck.}
\label{table:appendix_faster_trainig_aliased}
\end{table*}

\textbf{$3$D synthetic environments: } As discussed in Sec.\ref{sec:exp_aliased_room}, we need $M=4$ discrete bottlenecks for the \texttt{TDB} training to converge in $3$D synthetic environments. Table \ref{table:appendix_faster_trainig_agix} illustrates this by reporting \texttt{TrainingAfter1k}---the training accuracy after $1000$ training steps---for the different models reported in Table \ref{table:agix_room}, Appendix \ref{appendix:agix_environment}.

\begin{table*}[!ht]
\centering
\resizebox{0.45\textwidth}{!}{
\begin{tabular}{p{0.25\textwidth}p{0.25\textwidth} }
    \toprule
    Method & TrainingAfter1k $(\%)$ $\uparrow$ \\
    \toprule
    \toprule
    \texttt{TDB}$(S=1, M=1)$ & $41.40~(0.09)$ \\
    \texttt{TDB}$(S=1, M=4)$ & $99.72~(0.03)$\\
    \midrule
    \texttt{TDB}$(S=1, \enc, M=1)$ & $40.81~(0.93)$ \\
    \texttt{TDB}$(S=1, \enc, M=4)$ & $99.71~(0.04)$\\
    \midrule
    \texttt{TDB}$(S=3, M=1)$ & $28.05~(1.88)$\\
    \texttt{TDB}$(S=3, M=4)$ & $99.63~(0.06)$\\
    \bottomrule
\end{tabular}}
\caption{On the $3$D synthetic environments, training a \texttt{TDB} with $M=4$ discrete bottlenecks converges faster than training its counterpart with a single discrete bottleneck.}
\label{table:appendix_faster_trainig_agix}
\end{table*}

\textbf{GINC text dataset: } On the GINC dataset, we also evaluate the training accuracy of each model reported in Sec.\ref{sec:exp_icl} at every $1000$ training steps. Here, \texttt{TrainingAbove98} reports the first iteration when the training accuracy becomes higher than $98\%$ of the highest value it reaches during training.

\begin{table*}[!ht]
\centering
\resizebox{0.45\textwidth}{!}{
\begin{tabular}{p{0.25\textwidth}p{0.25\textwidth} }
    \toprule
    Method & TrainingAfter1k $(\%)$ $\uparrow$ \\
    \toprule
    \toprule
    \texttt{TDB}$(S=1, M=1)$ & $16000$ \\
    \texttt{TDB}$(S=1, M=4)$ & $6000$\\
    \midrule
    \texttt{TDB}$(S=3, M=1)$ & $13000$ \\
    \texttt{TDB}$(S=3, M=4)$ & $5000$\\
    \midrule
    \texttt{TDB}$(S=5, M=1)$ & $11000$\\
    \texttt{TDB}$(S=5, M=4)$ & $5000$\\
    \bottomrule
\end{tabular}}
\caption{On the GINC dataset, training a \texttt{TDB} with $M=4$ discrete bottlenecks converges faster than training its counterpart with a single discrete bottleneck.}
\label{table:appendix_faster_trainig_ginc}
\end{table*}

\bigskip

\newpage
\section{Do multiple discrete bottlenecks learn a disentangled representation?}\label{appendix:disentanglement}

\paragraph{Predicting a discrete bottleneck index given the other indices: } We propose herein a metric to test whether multiple discrete bottlenecks learn a disentangled representation of the observations.

If the discrete bottlenecks learn of a disentangled representation, then (a) they have to be independent (b) each one has to explain a distinct factor of the data \citep{carbonneau2020measuring}. Given a \texttt{TDB} with $M$ discrete bottlenecks, we propose to estimate (a), that is, whether the bottlenecks are independent. To do so, we train a logistic regression estimator to predict each bottleneck index given the other $M-1$ bottleneck indices. 

Given a sequence of observations $(x_1, \ldots, x_N)$, a \texttt{TDB} with $M$ discrete bottlenecks maps each observation $x_n$ to the latent codes $\left(s_n^{(1)}, \ldots, s_n^{(M)} \right)$ where $s_n^{(i)} =\phi^{(i)}(T_{2n-1}) \in \{ 1, \ldots, K\}$---see Sec.\ref{sec:mutiple_dbs}. 

We denote $e^{(i)}_n = \left(\mathbf{1}(s_n^{(i)}=k) \right)_{1\le k \le K}$ the one-hot encoding of $s_n^{(i)}$.

Let us now assume that we want to predict the $M$th bottleneck index given the first $M-1$ bottleneck indices.

We note $z = \left(e^{(1)}_n, \ldots, e^{(M-1)}_n \right) \in \{0, 1\}^{(M-1) K}$ and $y = e^{(M)}_n \in \{0, 1\}^{K}$, where we drop the dependency over $n$. We also introduce a matrix $W \in \mathbb{R}^{K \times (M-1)K}$. The logistic regression objective at timestep $n$ is:
\begin{equation*}
\mathcal{L}^{(M)}_{\text{disentanglement}}(n)
= - \sum_{i=1}^K
y_i ~ \log \left( \frac{\exp (W_i^T z)}{\sum_{j=1}^K \exp (W_j^T z) } \right)
\end{equation*}
We also define $\mathcal{L}^{(M)}_{\text{disentanglement}} = \sum_{n=1}^N \mathcal{L}^{(M)}_{\text{disentanglement}}(n)$. Note that we do not use an intercept term or a regularization term.

\paragraph{Training: }
Given a training and test set from Sec.\ref{sec:exp}, each consisting of $N_{\text{seq}}$ test sequences of length $N$, we first compute all the active bottleneck indices, which gives us two tensors, $\mathcal{T}_{\text{train}}$ and $\mathcal{T}_{\text{test}}$, each of dimensions $N_{\text{seq}} \times N \times M$. 

We train $M$ logistic regression estimators on $\mathcal{T}_{\text{train}}$: the $i$th estimator is trained to minimize $\mathcal{L}^{(i)}_{\text{disentanglement}}$, that is, to predict the $i$th bottleneck index given the other $M-1$. For training, we use Adam \citep{kingma2014adam} for $5,000$ iterations, with a learning rate of $0.001$ and a batch size of $32$.

\paragraph{Disentanglement metric:}
Our proposed disentanglement metric is the averaged test accuracy of the $M$ logistic regression estimators on the test tensor $\mathcal{T}_{\text{test}}$.

\paragraph{Results:}
In Sec.\ref{sec:exp}, we use $K=1000$ latent codes for each discrete bottleneck: if the bottleneck representations are independent, the test disentanglement accuracy would be $0.20\%$. However, all the test disentanglement accuracies reported in the main paper, are above $85\%$, which shows that multiple discrete bottlenecks learn highly redundant representations.

\newpage
\section{TDB with single-step objective merges nodes with identical neighbors}\label{appendix:merging_single_step}

In Sec.\ref{sec:intuition}, we make the argument that a \texttt{TDB} only trained to predict the next observation is not encouraged to disambiguate observations with identical neighbors in the ground truth (GT) room and, as a result, may merge their representations. We use this same argument in Sec.\ref{sec:objective_augmentation} to justify the need for augmenting the loss via either (a) a multi-step objective or (b) next encoding prediction.

\begin{figure*}[!h]
    \centering
    \includegraphics[height=0.31\textwidth]{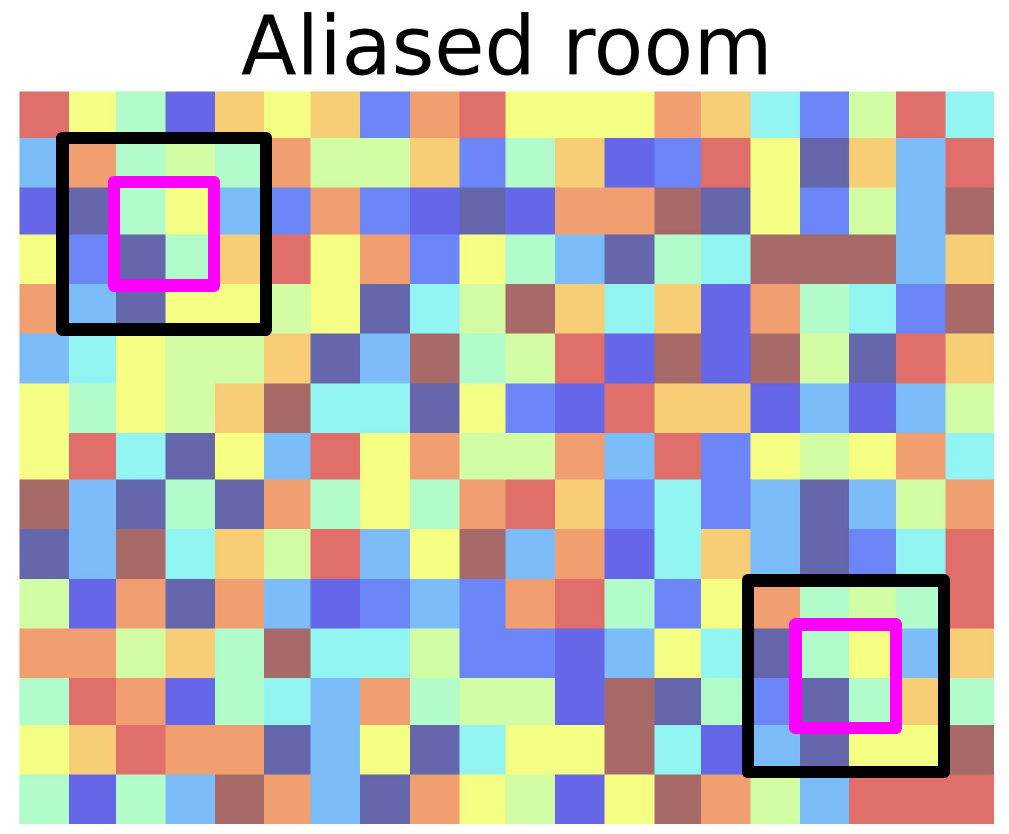}
    \hspace{0.02\textwidth}
    \includegraphics[height=0.31\textwidth]{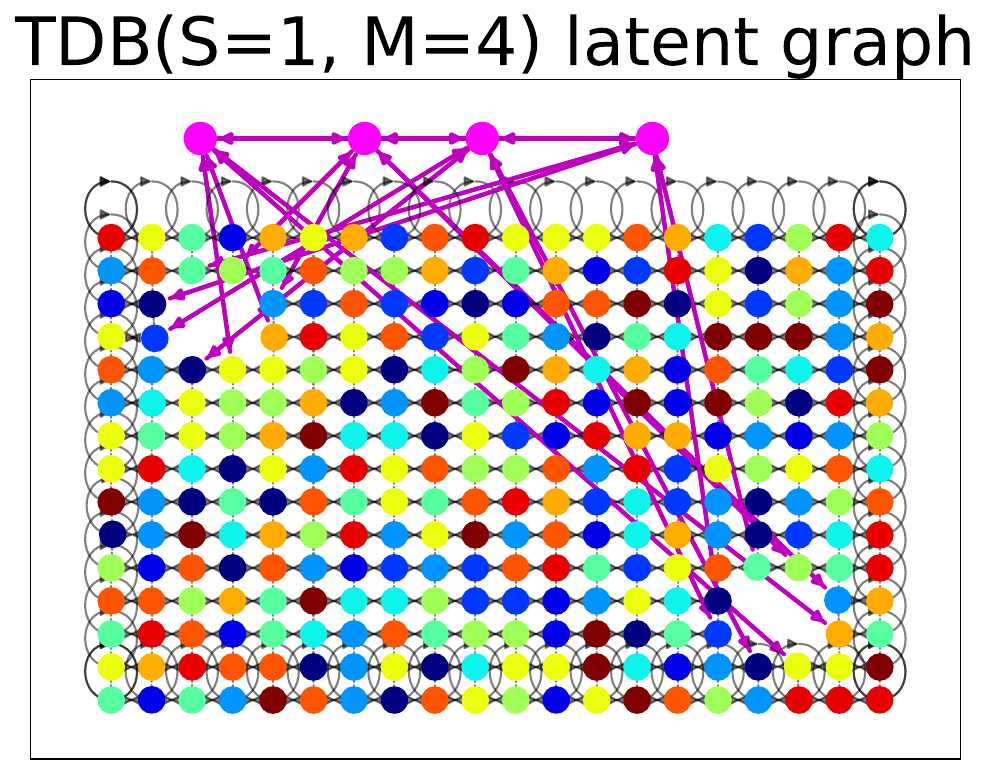}
    \caption{[Left] A 2D aliased room with $O=12$ unique observations: for each observation inside the $2\times2$ patches with fuchsia boundaries, all four actions (move up, down, left, right) lead to the same neighbors, regardless of whether the node is in the top-left $2\times2$ patch or the bottom-right one. [Right] The latent graph learned by a \texttt{TDB} with single step prediction fails at learning the ground truth dynamics. \texttt{TDB} learns the same representation for the two elements of each pair. This results in four merged nodes---colored in fuchsia---which are active at two spatial locations.  While these merged nodes do not affect test prediction performance, they introduce unrealistic shortcuts in the latent graph, which fool the external planner.}
    \label{fig:appendix_merging_single_step_aliased}
\end{figure*}

Fig.\ref{fig:appendix_merging_single_step_aliased} demonstrates the argument. We consider a $2$D aliased GT room from Sec. \ref{sec:exp_aliased_room} with $O=12$ unique observations---the numerical results are presented in Table \ref{table:aliased_appendix}, Appendix \ref{appendix:aliased_environment_12}. Similar to Fig.\ref{fig:motivation}, the GT room contains a $4\times4$ patch with black borders repeated twice. Each $4\times4$ patch contains an inner $2\times2$ patch with fuchsia borders. For each node inside this inner patch, all four actions in the GT room (move up, down, left, and right) lead to the same neighbors, regardless of whether the node in the top-left $2\times2$ patch or the bottom-right one.

Fig.\ref{fig:appendix_merging_single_step_aliased}[right] shows the latent graph learned by a \texttt{TDB}$(S=1, M=4)$ with single step prediction and four discrete bottlenecks. Because this model is only trained to predict the next observation given the history, it learns the same representation for each pair of nodes with same location inside the inner $2\times2$ patch. This results in the presence of four \textit{merged nodes} in the latent graph---colored in fuchsia. When a merged node is active, the agent can (a) perfectly predict the next observation but (b) it cannot know which one of the two possible $2\times2$ patches it is in. In addition, the merged nodes introduce \textit{unrealistic shortcuts} in the learned latent graph: nodes that are far apart in the GT room are close in the latent graph. Because the learned cognitive map does not accurately model the ground truth dynamics, the external solver will propose shortest paths that are invalid in the GT room. As a consequence, as indicated in Table \ref{table:aliased_appendix}, Appendix \ref{appendix:aliased_environment_12}, \texttt{TDB}$(S=1, M=4)$ can only solve  $59.37\% (\pm3.93\%)$ of the path planning problems.

\bigskip

\newpage

\textbf{For aliased rooms with 4 colors:}
When $O=4$, a higher frequency of observations have identical neighbors. The learned latent graph merges these observations, which introduces a larger number of unrealistic shortcuts. This explains the poor performance of \texttt{TDB}($S=1, M=1$) in Table \ref{table:aliased}, which only improves $6.61\% (\pm 0.53\%)$ of the time over the fallback path. Because of this inherent model failure, \texttt{TDB}($S=1, M=4$) similarly only improves over the fallback path $6.21\% (\pm 0.62\%)$ of the time. In fact, \texttt{TDB} is only able to solve the easiest path planning problems, for which the optimal shortest path consists of only a small number of actions. Indeed, the average length of the optimal shortest paths solved by \texttt{TDB}($S=1, M=4$) is $2.15~(\pm 0.07)$ while the average length of the optimal shortest paths over all the path planning problems is $10.80~(\pm 0.08)$.

\bigskip

\bigskip

\begin{figure*}[!h]
    \centering
    \includegraphics[height=0.31\textwidth]{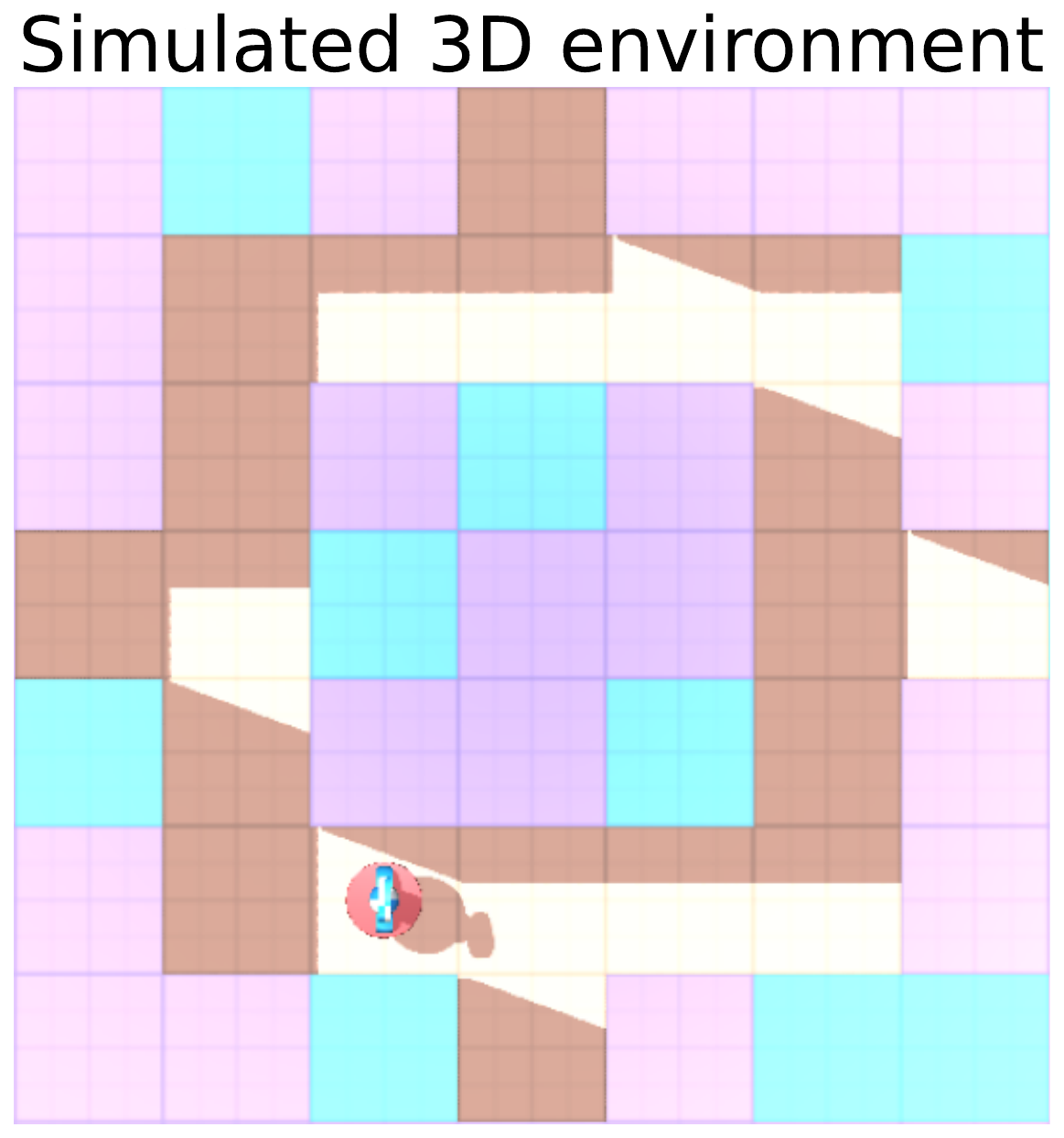}
    \hspace{0.02\textwidth}
    \includegraphics[height=0.31\textwidth]{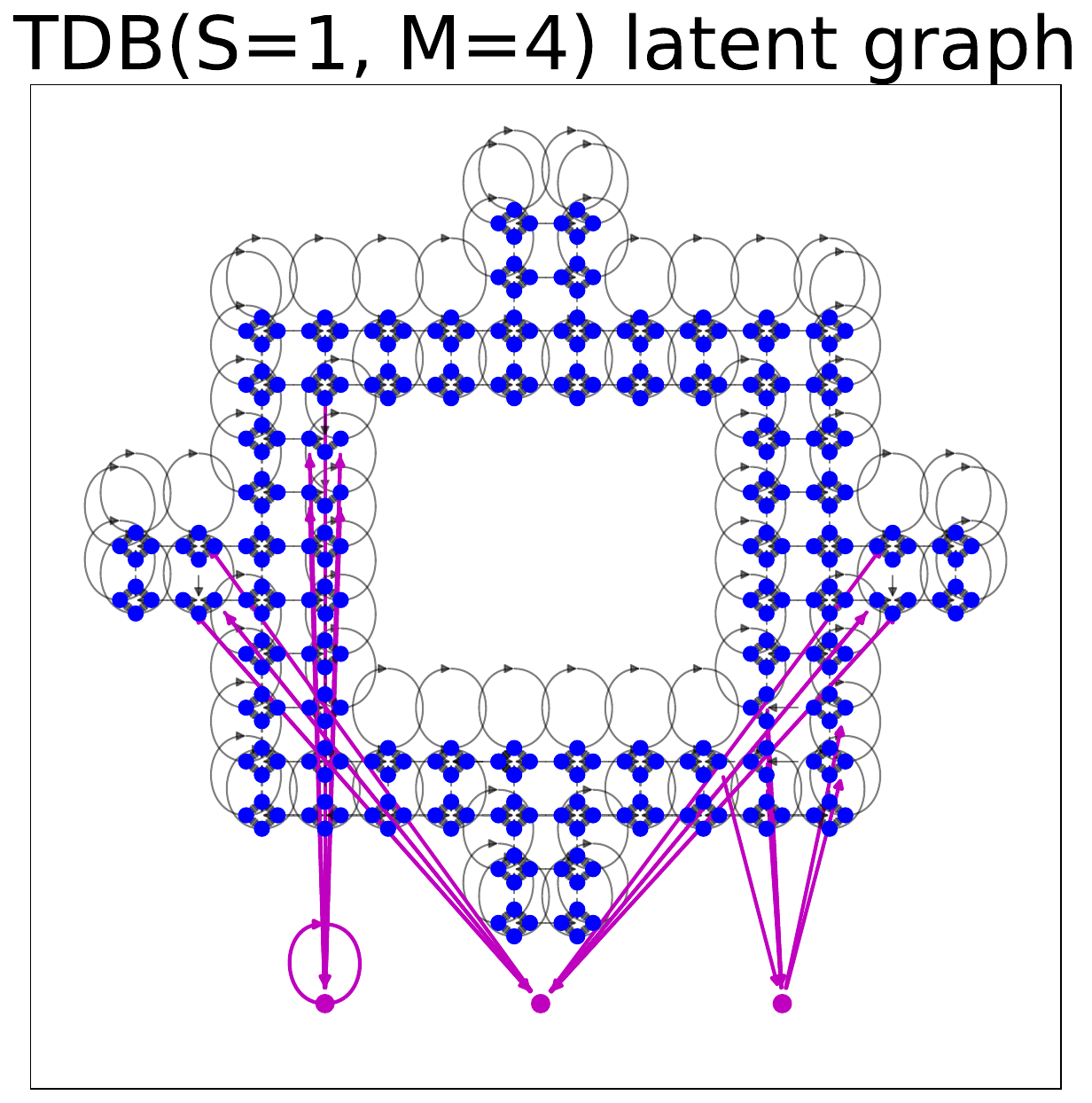}
    \hspace{0.02\textwidth}
    \includegraphics[height=0.31\textwidth]{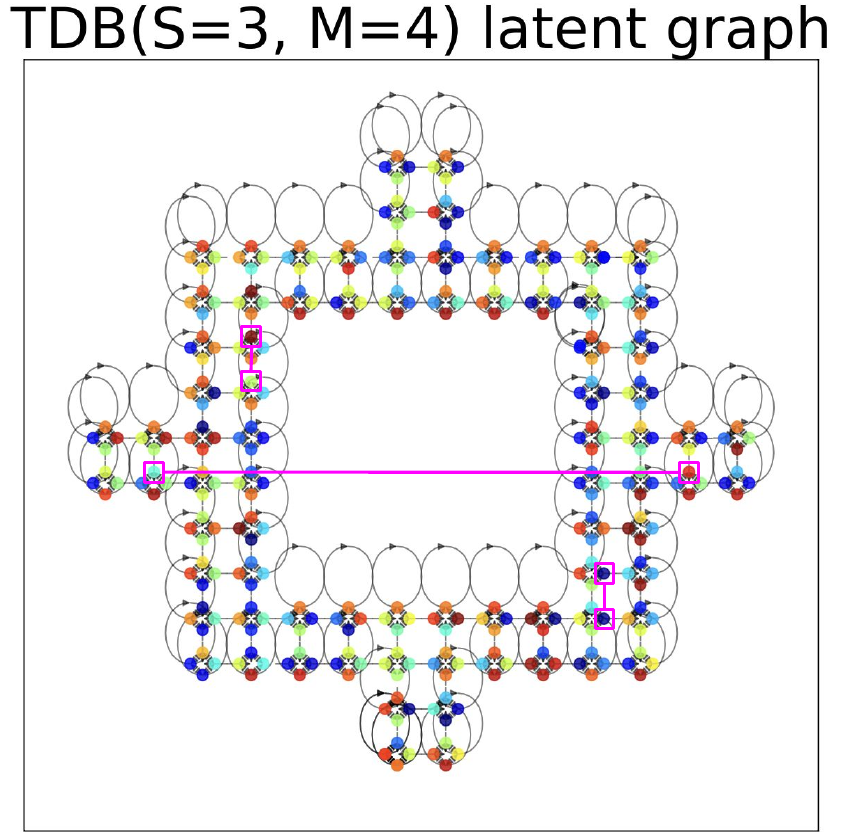}
    \caption{[Left] Top-down view of a $3$D simulated rooms used in Sec.\ref{sec:exp_agix}. [Middle] Latent graph learned with a \texttt{TDB}$(S=1, M=4)$ with single-step objective: each pair of observations with identical neighbors are represented with the same node, which introduces unrealistic shortcuts. [Right] Color-coded latent graph learned with a multi-step objective.}
    \label{fig:appendix_merging_single_step_agix}
\end{figure*}

\textbf{For 3D synthetic environments:}
Fig.\ref{fig:appendix_merging_single_step_agix}[middle] shows that, for a $3$D environment from Section \ref{sec:exp_agix}, the cognitive map learned with a \texttt{TDB}$(S=1, M=4)$ merges three pairs of nodes. These merged nodes introduce unrealistic shortcuts on the learned latent graph: as a result, \texttt{TDB}$(S=1, M=4)$ fails at path planning and only improves $65.30\% (\pm 4.24 \pm\%)$ of the time over the fallback path.

Fig.\ref{fig:appendix_merging_single_step_agix}[right] shows the latent graph learned when the objective is augmented to predict the next three steps, and highlights the three pairs of nodes merged in the middle image. Note that each node is mapped to the observation with higher empirical frequency when this node is active. As before, we see that the merged nodes have identical neighbors\footnote{Because there are $128$ clusters, the reader cannot precisely infer the observation from the colors: we looked at the graph indices to make sure that the neighbors of the merged nodes are identical.}.

\bigskip

\newpage
\section{Accelerating the graph edit distance}\label{appendix:ged}

The graph edit distance (\texttt{GED}) \citep{sanfeliu1983distance} between two graphs $\mathcal{G}_1$ and $\mathcal{G}_2$ is a graph similarity measure defined as minimum cost of edit path---i.e. sequence of node and edge edit operations---to transform the graph $\mathcal{G}_1$ into a graph isomorphic to $\mathcal{G}_2$.

Computing the exact \texttt{GED} is NP-complete. 
\newline
We therefore rely on the \texttt{\small{networkx.optimize\_graph\_edit\_distance}} \citep{hagberg2008exploring} function to return a sequence of approximations. This method accepts as argument a function to indicate which nodes (resp. edges) of $\mathcal{G}_1$ and $\mathcal{G}_2$ should be considered equal during matching. 

To accelerate \texttt{GED}, we say that a node $n_1$ of $\mathcal{G}_1$ and a node $n_2$ of $\mathcal{G}_2$ have to be considered equal --- and we note $n_1 \approx n_2$ if the ground truth spatial positions with higher empirical frequency when $n_1$ (resp. $n_2$) is active are the same. Similarly, we say that an edge of $\mathcal{G}_1$ connecting the nodes $(n^s_1, n^t_1)$ and an edge  of $\mathcal{G}_2$ connecting the nodes $(n^s_2, n^t_2)$ are considered to be equal if either (a) $n_1^s \approx n_2^s$ and $n_1^t \approx n_2^t$ or (b) $n_1^s \approx n_2^t$ and $n_1^t \approx n_2^s$.

We use a timeout of $900$s to compute \texttt{GED} in all the experiments but the ICL one, for which we use $20$s (due to the large number of normalized graph edit distances being computed).

\newpage
\section{Additional materials for aliased cube and for $2$D aliased room cubes}\label{appendix:aliased_environment}

\subsection{Table of results for non-Euclidean aliased cube}\label{appendix:aliased_cube}

Table \ref{table:aliased_appendix} reports the prediction and path planning metrics for an aliased cube with edge size $6$, similar to Fig. \ref{fig:agix_results}[bottom-left]. All the settings and parameters are the same as in Section \ref{sec:exp_aliased_room}. We average the results over $10$ experiments: each experiment considers a different cube.

\bigskip

\begin{table*}[!ht]
\centering
\resizebox{0.8\textwidth}{!}{
\begin{tabular}{p{0.23\textwidth}p{0.13\textwidth} p{0.15\textwidth} p{0.13\textwidth} p{0.13\textwidth}}
    \toprule
    Method & TestAccu ($\%$) $\uparrow$ & ImpFallback ($\%$) $\uparrow$ & RatioSP $\downarrow$ & NormGED $\downarrow$ \\
    \toprule
    \toprule
    Vanilla transformer &  $\mathbf{99.76}~(0.00)$  & $43.82~(1.17)$ & $22.74~(0.84)$ & ~~~~~~~~~~~~---\\
    \midrule
    Vanilla LSTM &  $\mathbf{99.76}~(0.00)$  & $43.82~(1.17)$ & $22.74~(0.84)$ & ~~~~~~~~~~~~---\\
    \midrule
    \texttt{TDB}$(S=1, \enc, M=1)$ &  $99.36~(0.06)$  & $99.07~(0.88)$ & $\mathbf{1.02}~(0.00)$ & $\mathbf{0.334}~(0.016)$\\
    \midrule
    \texttt{TDB}$(S=3, M=1)$ & $99.76~(0.00)$  & $\mathbf{99.05}~(0.23)$ & $1.03~(0.00)$ & $0.304~(0.011)$\\
    \bottomrule
\end{tabular}}
\caption{Results averaged over $10$ aliased cubes of edge size $6$. Our \texttt{TDB} with either multi-step objective or next encoding prediction (a) retains the nearly perfect test accuracy of vanilla sequence models (b) consistently solves the shortest paths problems when paired with an external solver---while both transformer or LSTM catastrophically fail---(c) learns cognitive maps almost isomorphic to the ground truth.
}
\label{table:aliased_cube}
\end{table*}

\newpage
\subsection{Table of results for 12 unique observations}\label{appendix:aliased_environment_12}

Table \ref{table:aliased_appendix} reports the prediction and path planning metrics for an aliased room with $O=12$ unique observations and global aliasing---as a $4\times4$ patch is repeated twice. All the settings and parameters are the same as in Section \ref{sec:exp_aliased_room}.

First, during training, a \texttt{TDB} with a single discrete bottleneck either (a) is stuck in a local optimum or (b) converges after a long time. As a consequence, contrary to Table \ref{table:aliased_appendix} which considers $O=4$, all the \texttt{TDB}s with a single bottleneck are here outperformed by their counterparts with four bottlenecks. \texttt{TDB}s with a single bottleneck also learn worse cognitive maps and solve a lower frequency of path planning problems.

Second, regardless of the number of bottleneck used, a \texttt{TDB} with single-step objective cannot disambiguate the global aliasing and introduces unrealistic shortcuts---see Appendix \ref{appendix:merging_single_step}---which result in a drop in planning metrics.

In contrast, \texttt{TDB}s with $M=4$ bottlenecks and either multi-step or next encoding objective (a) achieves nearly perfect prediction accuracy, (b) are able to almost consistently find the optimal shortest path, and (c) learn cognitive maps that recover the GT dynamics.

\bigskip

\begin{table*}[!ht]
\centering
\resizebox{0.8\textwidth}{!}{
\begin{tabular}{p{0.25\textwidth}p{0.13\textwidth} p{0.15\textwidth} p{0.13\textwidth} p{0.13\textwidth}}
    \toprule
    Method & TestAccu ($\%$) $\uparrow$ & ImpFallback ($\%$) $\uparrow$ & RatioSP $\downarrow$ & NormGED $\downarrow$ \\
    \toprule
    \toprule
    Vanilla transformer &  $\mathbf{99.75}~(0.00)$  & $29.89~(0.94)$ & $17.45~(0.57)$ & ~~~~~~~~~~~~---\\
    \midrule
    Vanilla LSTM &  $\mathbf{99.75}~(0.00)$  & $29.89~(0.94)$ & $17.45~(0.57)$ & ~~~~~~~~~~~~---\\
    \midrule
    \texttt{TDB}$(S=1, M=1)$ & $89.46~(2.48)$  & $17.31~(4.52)$ & $\mathbf{1.00}~(0.00)$ & $0.265~(0.053)$\\
    \texttt{TDB}$(S=1, M=4)$ & $\mathbf{99.75}~(0.00)$  & $59.37~(3.93)$ & $\mathbf{1.00}~(0.00)$ & $\mathbf{0.033}~(0.002)$\\
    \midrule
    \texttt{TDB}$(S=1, \enc, M=1)$ &  $92.57~(3.04)$  & $58.21~(10.98)$ & $\mathbf{1.00}~(0.00)$ & $0.145~(0.038)$\\
    \texttt{TDB}$(S=1, \enc, M=4)$ & $99.71~(0.00)$  & $96.58~(2.13)$ & $1.01~(0.00)$ & $0.157~(0.016)$\\
    \midrule
    \texttt{TDB}$(S=3, M=1)$ & $99.33~(0.01)$  & $97.09~(2.06)$ & $\mathbf{1.00}~(0.00)$ & $0.157~(0.010)$\\
    \texttt{TDB}$(S=3, M=4)$ & $99.73~(0.01)$  & $\mathbf{99.75}~(0.14)$ & $1.01~(0.00)$ & $0.160~(0.017)$\\
    \bottomrule
\end{tabular}}
\caption{
Prediction and path planning metrics averaged over $10$ aliased $15\times 20$ rooms with $O=12$ observations. Training a \texttt{TDB} single bottleneck ($M=1$) is slow and sometimes does not converge. A \texttt{TDB} with single-step objective cannot disambiguate global aliasing, fails at recovering the ground truth dynamics. \texttt{TDB} with multiple discrete bottlenecks ($M=4$) outperform their single bottleneck counterpart ($M=1$) for both prediction and path planning performance.
}
\label{table:aliased_appendix}
\end{table*}

\newpage
\subsection{The agent can locate itself after a short context} \label{appendix:aliased_environment_4}

\begin{figure*}[!h]
    \centering
    \includegraphics[width=0.45\textwidth]{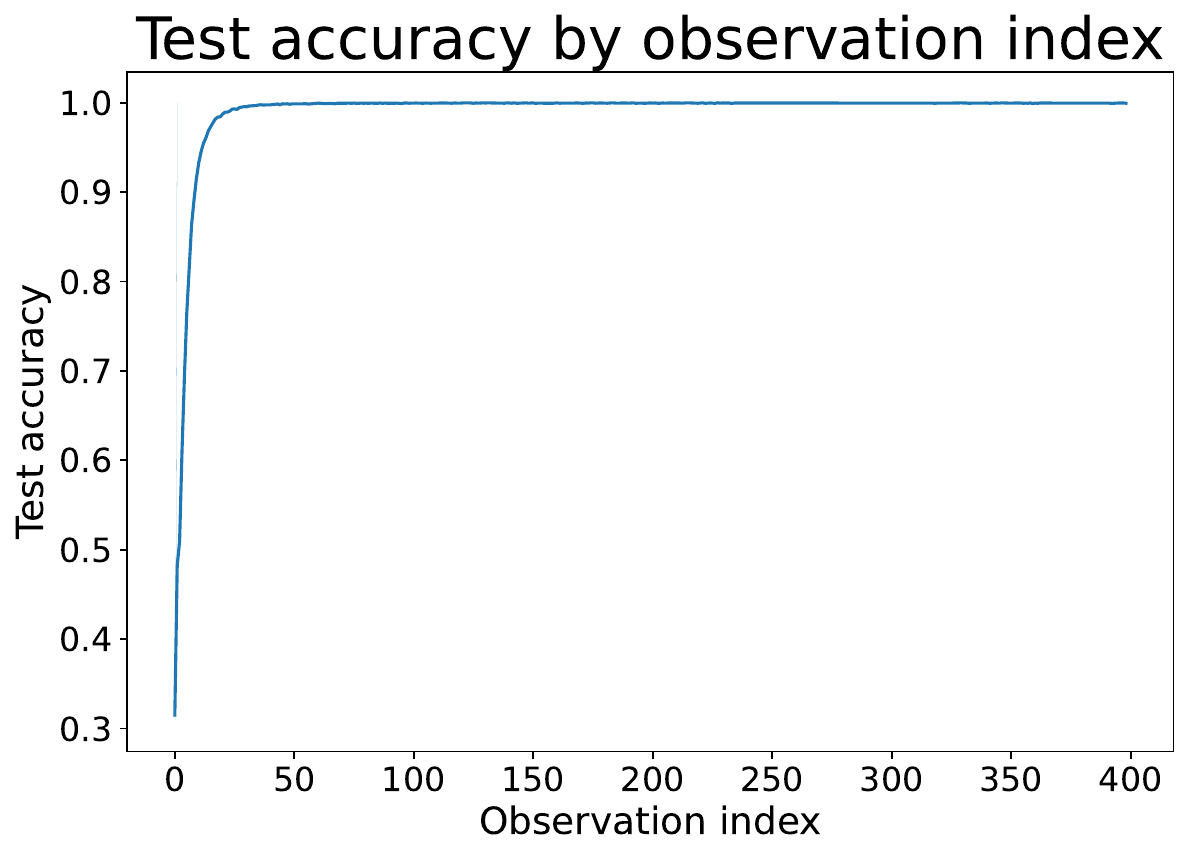}
    \caption{After a short context of $30$ observations, \texttt{TDB}$(S=3, M=4)$ predicts the next observations $99.50\%$ of the time.}
    \label{fig:appendix_prediction_accuracy_by_obs}
\end{figure*}

\paragraph{Test accuracy by timestep: }
Fig.\ref{fig:appendix_prediction_accuracy_by_obs} shows the average prediction accuracy as a function of the observation index for our best \texttt{TDB}$(S=3, M=1)$, reported in Sec.\ref{sec:exp_aliased_room} for the $2$D aliased room with $O=4$ observations. 

Because of aliasing, at the start of a test random walk, the model cannot know its location in the $2$D room: the prediction accuracy for the first timesteps is low. The agent then quickly locates itself and test accuracy rapidly increases: it is above $99.50\%$ after only $30$ observations.

\bigskip

\begin{figure*}[!h]
    \centering
    \includegraphics[width=0.5\textwidth]{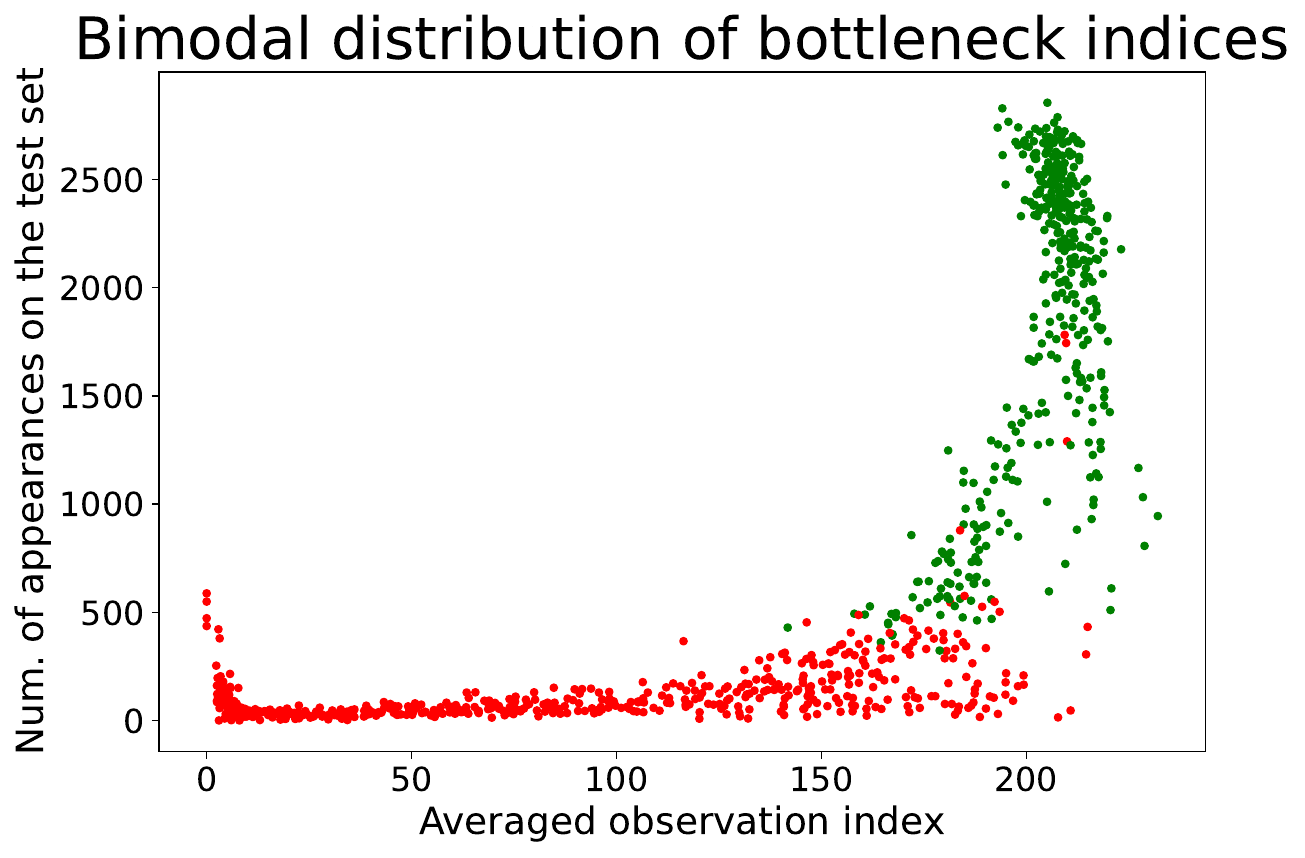}
    \caption{A first cluster of latent indices, in red, has low appearance counts and models the early agent's uncertainty in the test sequences. A second cluster, in green, has high appearance counts and represents the agent's confidence when it knows its location.}
    \label{fig:appendix_counts_vs_obs}
\end{figure*}

\textbf{Some latent codes model early uncertainty while others represent late confidence: }
Fig.\ref{fig:appendix_counts_vs_obs} displays, for each latent code active on the test data of a \texttt{TDB}$(S=3, M=1)$ (a) the average timestep at which this latent code is active---on the horizontal axis---and (b) its number of appearances on the entire test data---on the vertical axis. In addition, each bottleneck is colored in green (resp. red) to indicate that it is retained (resp. discarded) when we threshold the empirical count tensor $C$ to build the cognitive map---as detailed in Sec.\ref{sec:world_model}.

The figure displays two clusters, which highlight the two roles played by the latent codes. The first cluster represents a large number of codes with low appearance counts, which appear early in the test sequences. These codes model the agent's uncertainty when it does not know its location. 

The latent codes in the second cluster have a high number of appearances and appear later in the test sequences. They represent the agent's high confidence when it knows its locations. In addition, because we use a high count threshold $t_{\text{ratio}}=0.1$ (see Sec.\ref{sec:world_model}) to build the cognitive map, only the transitions between the latent codes of this second cluster are used to build the graph. As a result, a vast majority of the nodes of the second (resp. first) cluster are green (resp. red). Consequently, our third clustering step, detailed in Appendix \ref{sec:appendix_cluster}, maps the discarded (red) latent indices to their closest retained (green) latent index.

\bigskip

\textbf{The agent can ``teleportate'': }
We finally study how \texttt{TDB}$(S=3, M=1)$ moves in the learned latent graph. In particular, we say that the agent ``teleportates'' when the node of the transition graph where it estimates its location at the $n$th timestep is not a neighbor of its estimated position at the $n-1$th timestep. That is, teleportation means that the shortest path between two consecutive nodes is larger than one.

\begin{figure*}[!h]
    \centering
    \includegraphics[width=0.48\textwidth]{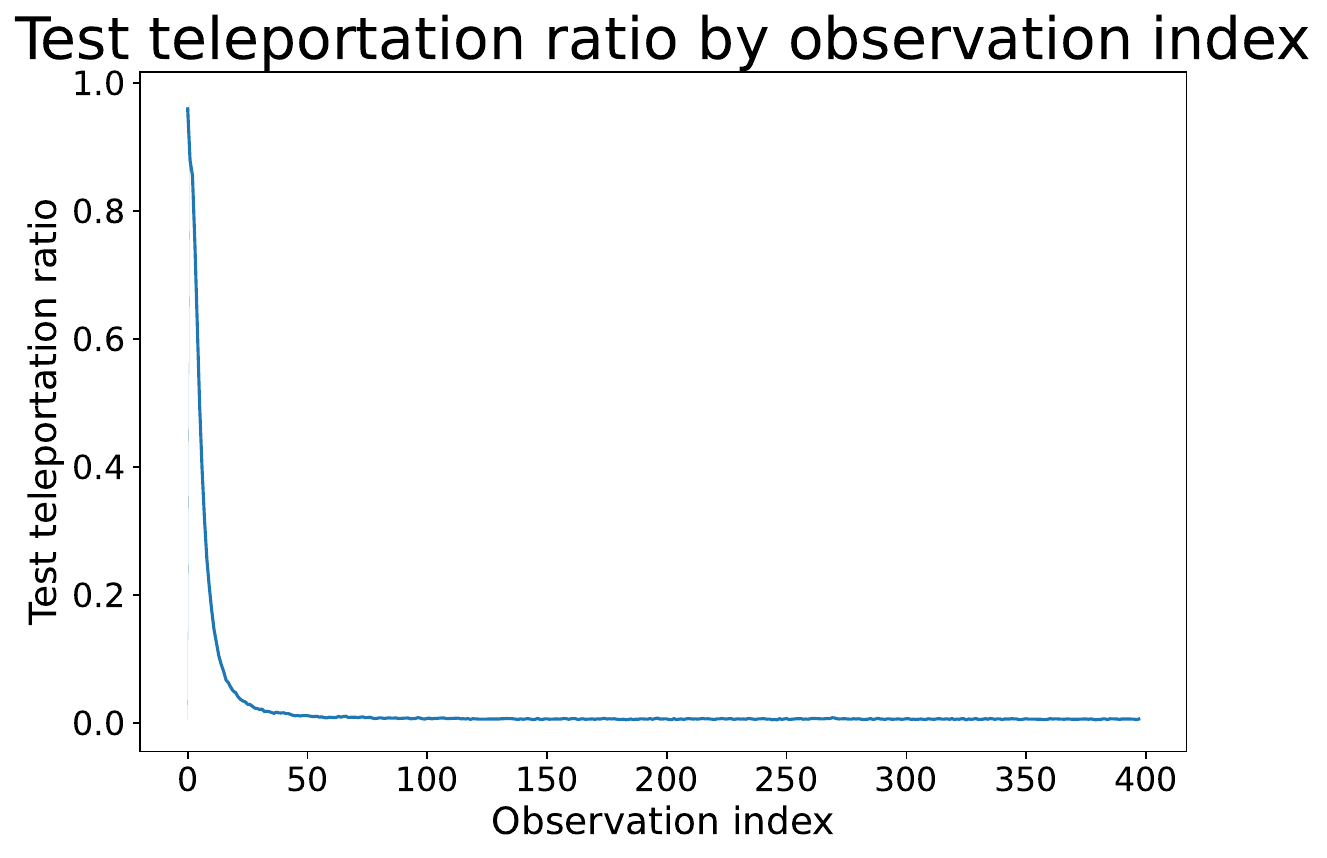}
    \vspace{0.02\textwidth}
    \includegraphics[width=0.45\textwidth]{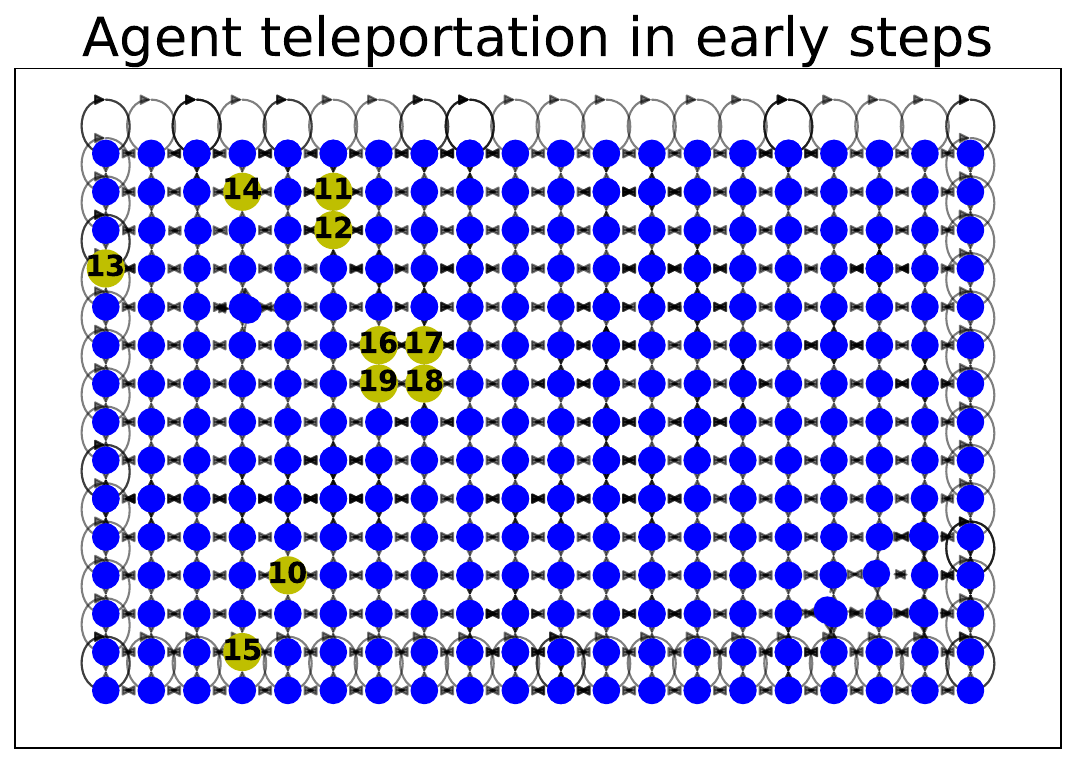}
    \caption{[Left] The agent frequently teleportates, i.e., moves between nodes that are not neighbors early in the test sequences. Teleportation rate drops below $5\%$ after $20$ observations, and below $1\%$ after $100$ observations. [Right] Despite perfectly predicting the next observation between indices $10$ and $20$, the agent also teleportates five steps out of ten. Here, teleportation is explained by incorrect clustering when we build the cognitive map.}
    \label{fig:appendix_teleportation}
\end{figure*}

Fig.\ref{fig:appendix_teleportation} looks at the average number of teleportation as a function of the observation index. Around the $20$th observation, the agent still teleportates $5\%$ of the time. As the agent becomes confident about its location, the teleportation rate drops below $1\%$ after $100$ observations and stays around $0.75\%$. 

Interestingly, Fig.\ref{fig:appendix_prediction_accuracy_by_obs}[right] shows a subsequence of $10$ observations $(x_{10},\ldots,x_{19})$ on a test random walks, with perfect next observation accuracy. Nonetheless,  the agent still ``teleportates'' five times out of ten. 

Here, teleportation is an artifact of the clustering procedures that we use when we build the cognitive map---see Appendix. \ref{sec:appendix_cluster}. For instance, our third clustering step may incorrectly map some discarded bottleneck indices to their closest bottleneck index. As a consequence, when the agent activates a discarded bottleneck index, it may be mapped to an invalid room location. If we use a lower threshold $t_{\text{ratio}}$, then teleportations should disappear but the learned latent graph would not accurately model the environment's dynamics.

\newpage
\section{Additional materials for simulated 3D environment}\label{appendix:agix_environment}

\paragraph{Table of results: } We first present the table of results for the simulated $3$D environment experiments in Sec.\ref{sec:exp_agix}.

\begin{table*}[!h]
\centering
\resizebox{0.8\textwidth}{!}{
\begin{tabular}{p{0.25\textwidth}p{0.13\textwidth} p{0.15\textwidth} p{0.13\textwidth} p{0.13\textwidth}}
    \toprule
    Method & TestAccu ($\%$) $\uparrow$ & ImpFallback ($\%$) $\uparrow$ & RatioSP $\downarrow$ & NormGED $\downarrow$ \\
    \toprule
    \toprule
    Vanilla transformer & $\mathbf{99.80}~(0.00)$ &  $26.60~(0.58)$ & $16.38~(0.87)$ & ~~~~~~~~~~~~---\\
    \midrule
    Vanilla LSTM & $99.79~(0.00)$ &  $26.85~(0.58)$ & $16.38~(0.87)$ & ~~~~~~~~~~~~---\\
    \midrule
    \texttt{TDB}$(S=1, \enc, M=1)$ &  $72.88~(2.86)$  & $2.85~(1.33)$ & $1.32~(0.21)$ & $0.733~(0.082)$\\
    \midrule
     \texttt{TDB}$(S=3, M=1)$ & $58.51~(1.43)$  & $1.35~(0.27)$ & $1.35~(0.13)$ & $0.946~(0.007)$\\
    \midrule
    \texttt{TDB}$(S=1, \enc, M=4)$ & $99.79~(0.01)$ &  $97.55~(0.42)$ & $1.03~(0.01)$ & $\mathbf{0.005}~(0.001)$ \\
    \midrule
    \texttt{TDB}$(S=3, M=4)$ & $99.79~(0.00)$ &  $\mathbf{99.95}~(0.05)$ & $\mathbf{1.02}~(0.00)$ & $0.017~(0.004)$ \\
    \bottomrule
\end{tabular}}
\caption{Metrics averaged over $10$ simulated $3$D environments. Training a \texttt{TDB} with a single discrete bottleneck does not converge. In contrast, a \texttt{TDB} with multiple discrete bottlenecks can almost perfectly (a) predict the next observation given history, (b) solve shortest paths problems and (c) learn cognitive maps nearly isomorphic to the ground truth map.}
\label{table:agix_room}
\end{table*}

\bigskip

\bigskip

\begin{figure*}[!h]
    \centering
    \includegraphics[width=.95\textwidth]{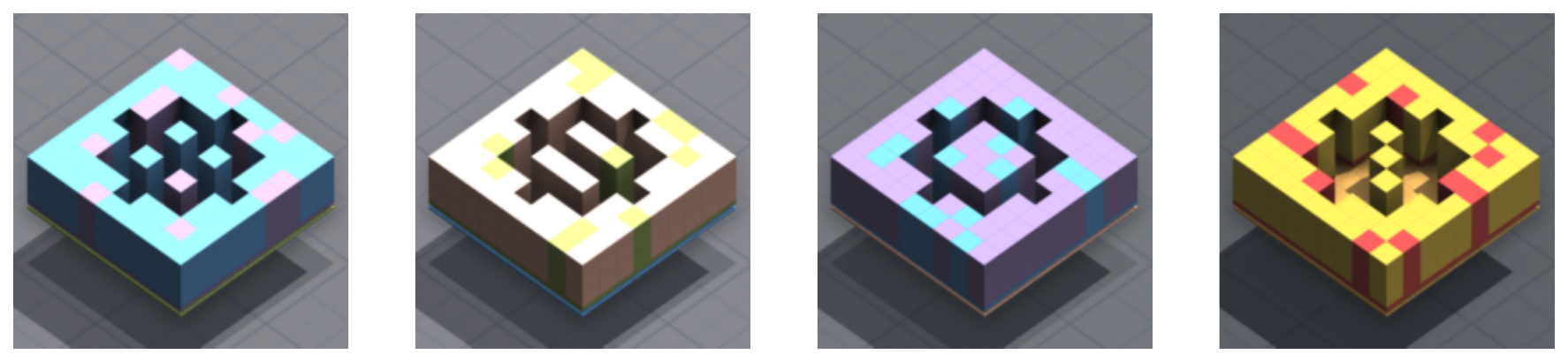}
    \includegraphics[width=0.21\textwidth]{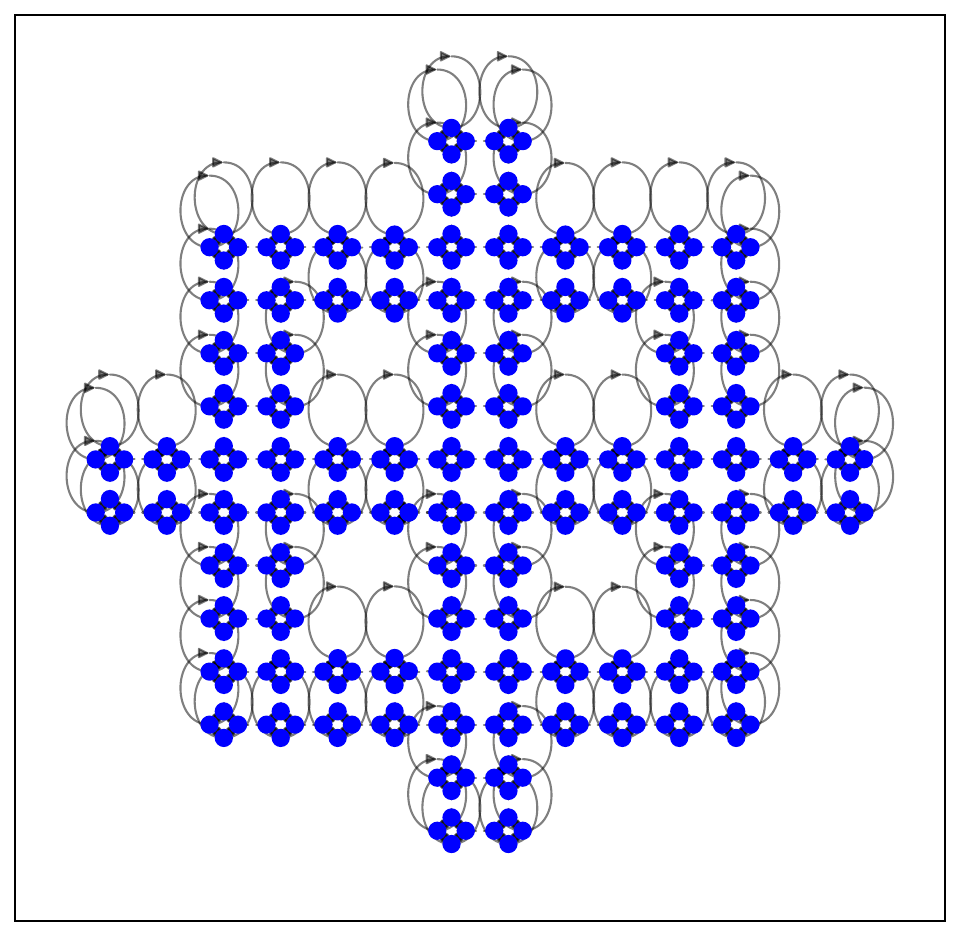}
    \hspace{0.02\textwidth}
    \includegraphics[width=0.21\textwidth]{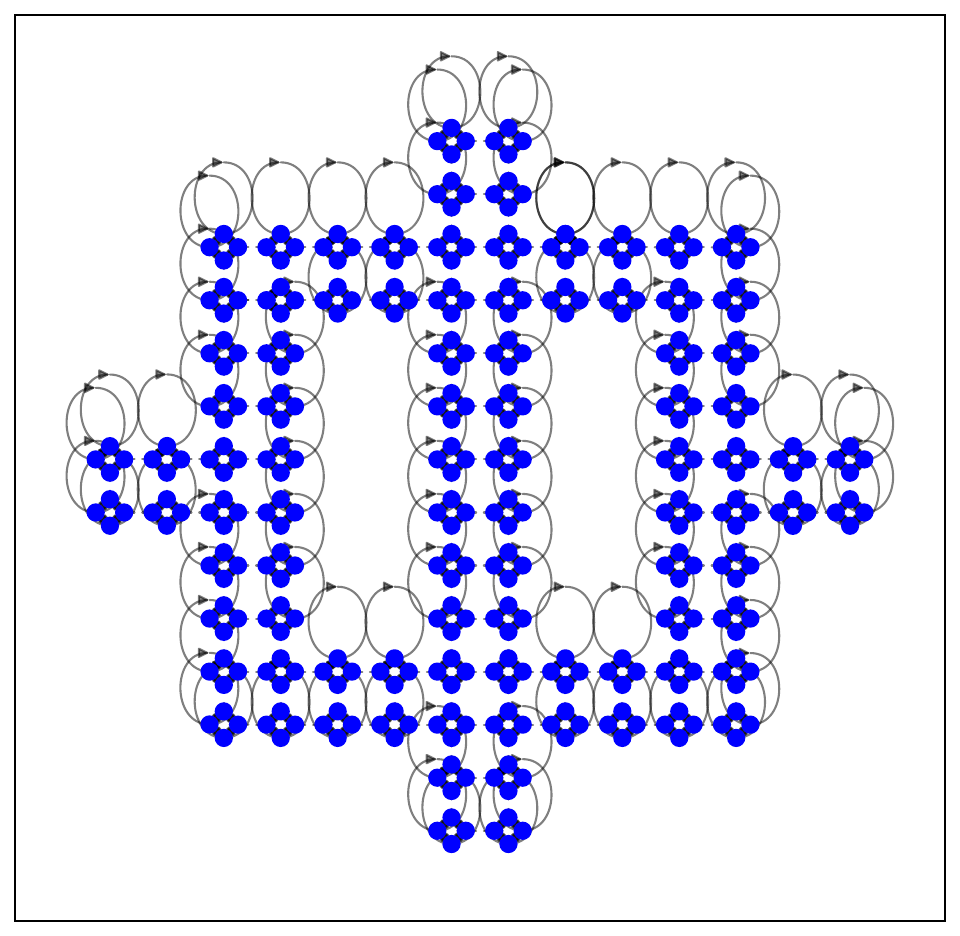}
    \hspace{0.02\textwidth}
    \includegraphics[width=0.21\textwidth]{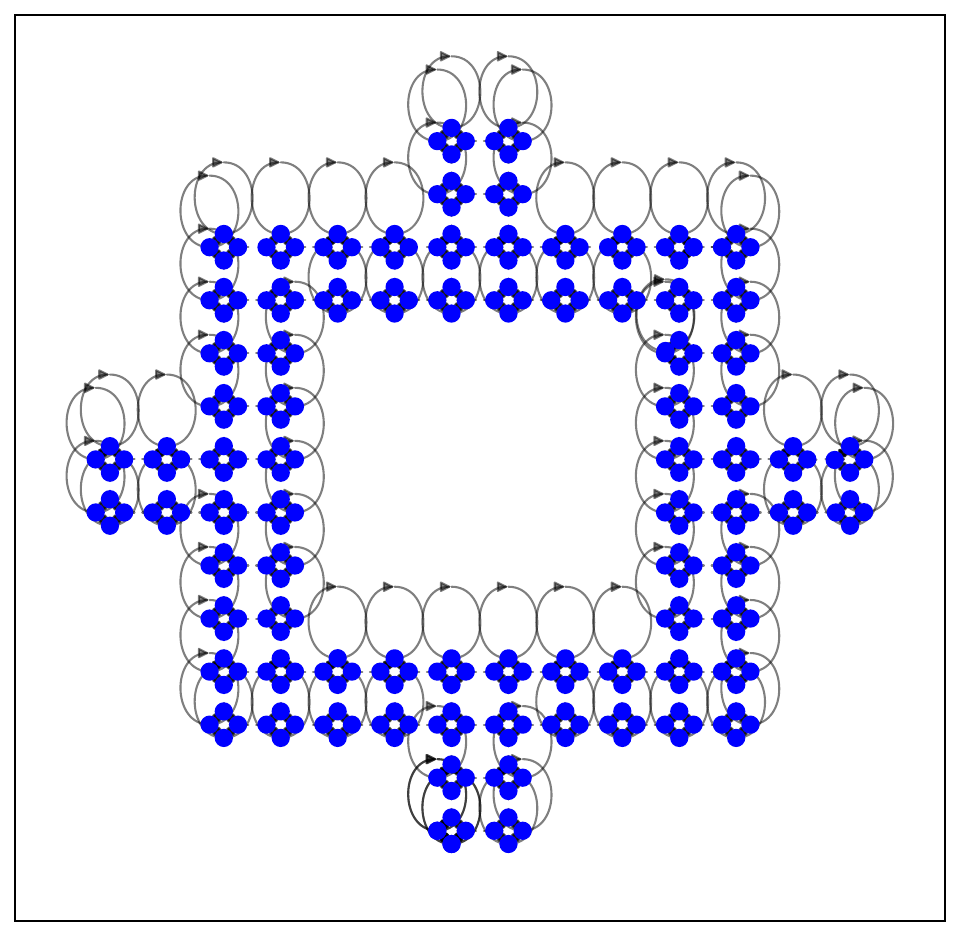}
    \hspace{0.02\textwidth}
    \includegraphics[width=0.21\textwidth]{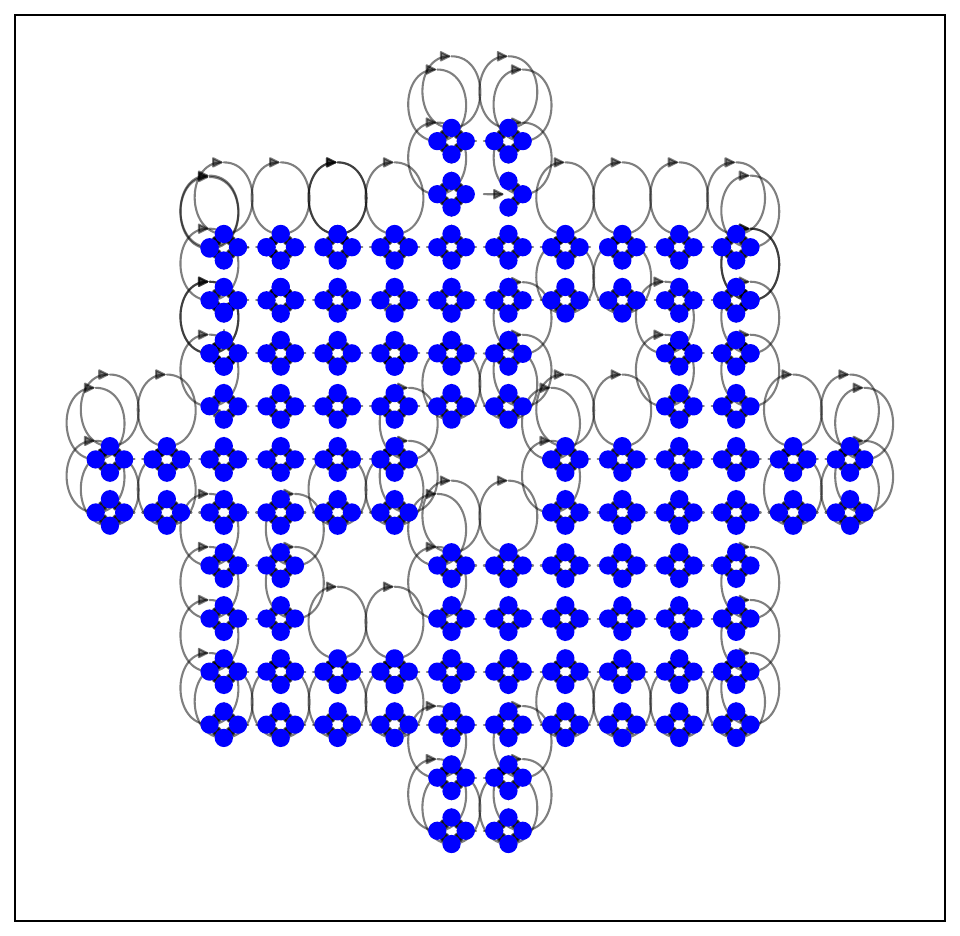}
    \caption{[Top] Isometric views of four $3$D simulated rooms used in Sec.\ref{sec:exp_agix}. [Bottom] Corresponding cognitive maps learned with a TDB($S=3, M=4$): each location in the room is represented by four nodes in the latent graph, corresponding to the four possible heading directions of the agent.}
    \label{fig:agix_appendix}
\end{figure*}

\paragraph{Latent graphs visualizations:}
Second, Fig.\ref{fig:agix_appendix} presents the isometric views of four $3$D simulated rooms, as well as the respective cognitive maps learned with a TDB($S=3, M=4$).

\newpage
\section{Additional materials for GINC dataset}\label{appendix:ginc}

Table \ref{table:appendix_ginc} presents numerical results associated with Fig.\ref{fig:ginc_results}[left].

\begin{table*}[!ht]
\footnotesize
\centering
\resizebox{.98\textwidth}{!}{
\begin{tabular}{p{0.13\textwidth}p{0.16\textwidth} p{0.16\textwidth} p{0.16\textwidth} p{0.18\textwidth} p{0.18\textwidth}}
\toprule
Context length & Num. of examples & LSTM & Transformer  & TDB$(S=3, M=4)$  & TDB$(S=5, M=1)$ \\
\toprule
\toprule
$3$ & $0$ & $0.426~(0.01)$ & $\mathbf{0.502}~(0.01)$ & $0.426~(0.01)$ & $0.284~(0.009)$ \\ 
& $1$ & $0.369~(0.01)$ & $\mathbf{0.488}~(0.01)$ & $0.464~(0.01)$ & $0.369~(0.01)$ \\ 
& $2$ & $0.387~(0.01)$ & $0.538~(0.01)$ & $\mathbf{0.554}~(0.01)$ & $0.443~(0.01)$ \\ 
& $4$ & $0.394~(0.01)$ & $0.553~(0.01)$ & $\mathbf{0.588}~(0.01)$ & $0.526~(0.01)$ \\ 
& $8$ & $0.386~(0.01)$ & $0.595~(0.01)$ & $\mathbf{0.618}~(0.01)$ & $0.549~(0.01)$ \\ 
& $16$ & $0.390~(0.01)$ & $0.631~(0.01)$ & $\mathbf{0.655}~(0.01)$ & $0.574~(0.01)$ \\ 
& $32$ & $0.380~(0.01)$ & $0.633~(0.01)$ & $\mathbf{0.670}~(0.009)$ & $0.618~(0.01)$ \\ 
& $64$ & $0.377~(0.01)$ & $0.651~(0.01)$ & $\mathbf{0.676}~(0.009)$ & $0.646~(0.01)$ \\ 
\midrule
$5$ & $0$ & $0.692~(0.009)$ & $\mathbf{0.805}~(0.008)$ & $0.777~(0.008)$ & $0.656~(0.01)$ \\ 
& $1$ & $0.742~(0.009)$ & $0.822~(0.008)$ & $\mathbf{0.853}~(0.007)$ & $0.768~(0.008)$ \\
& $2$ & $0.754~(0.009)$ & $0.831~(0.008)$ & $\mathbf{0.857}~(0.007)$ & $0.802~(0.008)$ \\
& $4$ & $0.749~(0.009)$ & $0.841~(0.007)$ & $\mathbf{0.905}~(0.006)$ & $0.872~(0.007)$ \\
& $8$ & $0.761~(0.009)$ & $0.847~(0.007)$ & $\mathbf{0.901}~(0.006)$ & $0.882~(0.006)$ \\
& $16$ & $0.756~(0.009)$ & $0.871~(0.007)$ & $\mathbf{0.922}~(0.005)$ & $0.908~(0.006)$ \\ 
& $32$ & $0.767~(0.008)$ & $0.871~(0.007)$ & $\mathbf{0.926}~(0.005)$ & $0.915~(0.006)$ \\ 
& $64$ & $0.758~(0.009)$ & $0.88~(0.006)$ & $\mathbf{0.924}~(0.005)$ & $0.917~(0.006)$ \\ 
\midrule
$8$ & $0$ & $0.849~(0.007)$ & $0.881~(0.006)$ & $\mathbf{0.895}~(0.006)$ & $0.816~(0.008)$ \\
& $1$ & $0.869~(0.007)$ & $0.888~(0.006)$ & $\mathbf{0.930}~(0.005)$ & $0.899~(0.006)$ \\ 
& $2$ & $0.869~(0.007)$ & $0.886~(0.006)$ & $\mathbf{0.930}~(0.005)$ & $0.917~(0.006)$ \\ 
& $4$ & $0.877~(0.007)$ & $0.897~(0.006)$ & $0.933~(0.005)$ & $\mathbf{0.936}~(0.005)$ \\
& $8$ & $0.873~(0.007)$ & $0.897~(0.006)$ & $0.935~(0.005)$ & $\mathbf{0.940}~(0.005)$ \\ 
& $16$ & $0.875~(0.007)$ & $0.916~(0.006)$ & $0.942~(0.005)$ & $\mathbf{0.950}~(0.004)$ \\
& $32$ & $0.862~(0.007)$ & $0.909~(0.006)$ & $0.948~(0.004)$ & $\mathbf{0.956}~(0.004)$ \\ 
& $64$ & $0.883~(0.006)$ & $0.920~(0.005)$ & $0.952~(0.004)$ & $\mathbf{0.955}~(0.004)$ \\
\midrule
$10$ & $0$ & $0.878~(0.007)$ & $0.888~(0.006)$ & $\mathbf{0.915}~(0.006)$ & $0.858~(0.007)$ \\ 
& $1$ & $0.889~(0.006)$ & $0.898~(0.006)$ & $\mathbf{0.930}~(0.005)$ & $0.926~(0.005)$ \\
& $2$ & $0.878~(0.007)$ & $0.901~(0.006)$ & $\mathbf{0.938}~(0.005)$ & $0.935~(0.005)$ \\ 
& $4$ & $0.902~(0.006)$ & $0.904~(0.006)$ & $0.936~(0.005)$ & $\mathbf{0.945}~(0.005)$ \\ 
& $8$ & $0.881~(0.006)$ & $0.906~(0.006)$ & $0.947~(0.004)$ & $\mathbf{0.956}~(0.004)$ \\ 
& $16$ & $0.882~(0.006)$ & $0.908~(0.006)$ & $0.934~(0.005)$ & $\mathbf{0.947}~(0.004)$ \\ 
& $32$ & $0.889~(0.006)$ & $0.918~(0.005)$ & $0.946~(0.005)$ & $\mathbf{0.960}~(0.004)$ \\ 
& $64$ & $0.89~(0.006)$ & $0.911~(0.006)$ & $0.941~(0.005)$ & $\mathbf{0.956}~(0.004)$ \\ 
\bottomrule
\end{tabular}
}
\caption{In-context accuracy for vanilla sequence models and \texttt{TDB}s, for each each pair $(k, n)$ of context length $k$ and number of examples $n$ of the GINC test set. Our proposed \texttt{TDB}s consistently reach the highest in-context accuracies.}
\label{table:appendix_ginc}
\end{table*}

\newpage
\section{Additional materials for in-context learning}\label{appendix:icl}

\subsection{Defining the in-context learning problem}\label{appendix:icl_setup}
\paragraph{Preserving the room partition: }
We consider a first $2$D aliased room of size $6 \times 8$ with $O=4$ observations, $c_1, \ldots c_O$ such that $c_i \in \{1, \ldots, 30\}$. We represent this room by a matrix $R \in \{c_1, \ldots c_O\}^{6 \times 8}$, where, for $x\in \{1, \ldots, 6\}$ and $y\in \{1, \ldots, 8\}$, $R_{xy}$ is the room observation at the spatial position $(x, y)$.

We define the \textit{room partition} induced by the room colors as the partition $\mathcal{S}_1, \ldots, \mathcal{S}_O$ of the $2$D room spatial positions such that, for each element $\mathcal{S}_i$ of the partition, all the spatial positions in $\mathcal{S}_i$ have the same observation.

Formally, the room partition is defined as
$$
\bigcup_{i=1}^O \mathcal{S}_i = \{(x, y): ~ 1 \le x \le 6, ~ 1 \le y \le 8 \} 
~~~ \text{;} ~~~ \mathcal{S}_i \cap \mathcal{S}_j = \emptyset, ~\forall i \ne j 
~~~ \text{;} ~~~ \mathcal{S}_i \ne \emptyset, ~\forall i,
$$
and satisfies
$$
\forall i\le O, ~\forall (x, y) \in \mathcal{S}_i, ~R_{xy} = o_i.
$$

For each injective mapping $\phi: \{1, \ldots, O\}\to \{1, \ldots, 30\}$, we can build a new room $\tilde{R}$ with (a) observations $\phi(1), \ldots, \phi(O)$ and which (b) \textit{preserves the room partition}, by assigning all the $2$D spatial positions in $\mathcal{S}_i$ to the observation $\phi(i)$. That is
$$
\forall i\le O, ~\forall (x, y) \in \mathcal{S}_i, ~\tilde{R}_{xy} = \phi(i).
$$

The training sets in Sec.\ref{sec:exp_icl} contain at most $10$k such training rooms, each one preserving the room partition. Note that, for a given partition, the number of possible rooms which preserves the room partition is the number of injective mapping from $\{1, \ldots, O\}$ to $\{1, \ldots, 30\}$, which is $\prod_{i=0}^{O-1} 30 - i$.

\paragraph{Learning in base: } As discussed in the main text, given a sequence of observations $x=(x_1,\ldots,x_N)$ in a room defined as above, each model is trained to predict targets in the \textit{room-agnostic base}, $\tilde{x}_n \in \{1, \ldots,O\}$, such that $\tilde{x}_n=k$ iff. $\tilde{x}_n$ is the $k$th lowest observation seen between indices $1$ and $n$. 

In particular, let us define the permutation $i_1, \ldots, i_O$ of $1, \ldots, O$ such that $\phi(i_1) \le \ldots \le \phi(i_O)$. Let $n^*$ be such a large enough integer such that all the $O$ different room observations have been seen between $x_1$ and $x_{n^*}$. Then, for $n \ge n^*$, the target $\tilde{x}_n=k$ corresponds to the $k$th highest color of the room, which is equal to $\phi(i_k)$.

\subsection{Table of results} \label{appendix:icl_results}

Table \ref{table:appendix_icl} reports the numerical results displayed in Fig.~\ref{fig:icl_results}[left].

\begin{table*}[!ht]
\footnotesize
\centering
\resizebox{.98\textwidth}{!}{
\begin{tabular}{p{0.16\textwidth}p{0.16\textwidth} p{0.16\textwidth} p{0.16\textwidth} p{0.18\textwidth} p{0.2\textwidth}}
\toprule
Metric & Num. training rooms & LSTM & Transformer  & TDB$(S=3, M=4)$  & TDB$(S=1, \enc, M=4)$ \\
\toprule
\toprule
TestAccu ($\%$) $\uparrow$ & $200$ & $\mathbf{0.508}~(0.009)$ & $0.426~(0.005)$ & $0.38~(0.004)$ & $0.379~(0.002)$  \\ 
& $500$ & $\mathbf{0.741}~(0.039)$ & $0.595~(0.006)$ & $0.461~(0.009)$ & $0.481~(0.007)$  \\ 
& $1000$ & $\mathbf{0.899}~(0.006)$ & $0.864~(0.016)$ & $0.568~(0.015)$ & $0.58~(0.03)$  \\ 
& $2000$ & $0.919~(0.009)$ & $\mathbf{0.977}~(0.003)$ & $0.868~(0.03)$ & $0.8~(0.056)$  \\ 
& $5000$ & $0.848~(0.045)$ & $\mathbf{0.985}~(0.001)$ & $\mathbf{0.974}~(0.001)$ & $0.784~(0.091)$  \\ 
& $10000$ & $0.924~(0.007)$ & $\mathbf{0.985}~(0.001)$ & $0.977~(0.001)$ & $0.912~(0.059)$  \\ 
\midrule
ImpFallback ($\%$) $\uparrow$ & $200$ & $0.005~(0.001)$ & $0.002~(0.001)$ & $\mathbf{0.038}~(0.005)$ & $0.032~(0.004)$  \\ 
& $500$ & $0.148~(0.04)$ & $0.007~(0.002)$ & $0.05~(0.004)$ & $\mathbf{0.064}~(0.004)$  \\ 
& $1000$ & $0.382~(0.054)$ & $0.189~(0.032)$ & $0.115~(0.014)$ & $\mathbf{0.124}~(0.016)$  \\ 
& $2000$ & $0.447~(0.059)$ & $0.725~(0.024)$ & $\mathbf{0.690}~(0.067)$ & $0.495~(0.092)$  \\ 
& $5000$ & $0.312~(0.086)$ & $0.809~(0.009)$ & $\mathbf{0.964}~(0.011)$ & $0.605~(0.132)$  \\ 
& $10000$ & $0.305~(0.072)$ & $0.771~(0.011)$ & $\mathbf{0.970}~(0.009)$ & $0.782~(0.084)$  \\ 
\midrule
RatioSP $\downarrow$ & $200$ & $51.79~(20.83)$ & $34.79~(11.07)$ & $1.46~(0.09)$ & $\mathbf{1.33}~(0.06)$  \\ 
& $500$ & $29.73~(3.67)$ & $45.26~(23.87)$ & $\mathbf{1.17}~(0.05)$ & $1.27~(0.06)$  \\ 
& $1000$ & $25.52~(1.19)$ & $26.14~(1.34)$ & $\mathbf{1.18}~(0.04)$ & $1.20~(0.04)$  \\ 
& $2000$ & $27.01~(0.89)$ & $25.98~(0.83)$ & $\mathbf{1.07}~(0.03)$ & $1.11~(0.04)$  \\ 
& $5000$ & $24.79~(0.96)$ & $24.53~(0.63)$ & $\mathbf{1.01}~(0.0)$ & $1.06~(0.05)$  \\ 
& $10000$ & $24.27~(1.73)$ & $25.13~(0.45)$ & $\mathbf{1.01}~(0.0)$ & $\mathbf{1.01}~(0.0)$  \\ 
\midrule
NormGED $\downarrow$  & $200$ & ~~~~~~~~--- & ~~~~~~~~--- & $\mathbf{0.707}~(0.005)$ & $0.723~(0.006)$  \\ 
& $500$ & ~~~~~~~~--- & ~~~~~~~~--- & $\mathbf{0.658}~(0.006)$ & $0.660~(0.008)$  \\ 
& $1000$ & ~~~~~~~~--- & ~~~~~~~~--- & $\mathbf{0.579}~(0.018)$ & $0.602~(0.033)$  \\ 
& $2000$ & ~~~~~~~~--- & ~~~~~~~~--- & $\mathbf{0.164}~(0.042)$ & $0.29~(0.073)$  \\ 
& $5000$ & ~~~~~~~~--- & ~~~~~~~~--- & $\mathbf{0.019}~(0.003)$ & $0.294~(0.121)$  \\ 
& $10000$ & ~~~~~~~~--- & ~~~~~~~~--- & $\mathbf{0.017}~(0.002)$ & $0.118~(0.079)$  \\ 
\bottomrule
\end{tabular}
}
\caption{In-context metrics for vanilla sequence models and \texttt{TDB}s on the new test rooms. For a large number of training rooms, our best \texttt{TDB}$(S=3, M=4)$ (a) reaches nearly perfect in-context accuracy, close to the best performing transformer, (b) almost perfectly solves the in-context path planning problems, and (c) learns in-context latent graphs nearly isomorphic to the ground truth.}
\label{table:appendix_icl}
\end{table*}

First, we observe that for each model, in-context capacities emerge when we increase the number of rooms.
 
Second, while vanilla LSTM performs best for a small number of training rooms, it cannot reach near-perfect in-context accuracies when the number of training rooms increases---which a vanilla transformer can do. We believe that attention-like mechanisms are important for in-context capacities to fully emerge in novel test rooms.

Third, as observed in other experiments, vanilla sequence models cannot solve path planning problems. Indeed, for a large number of training rooms, a transformer improves at best $77\%$ of the times over the fallback path. However, when it does so, it finds paths that are on average $25$ times longer than the optimal paths. In contrast, our best \texttt{TDB}$(S=3, M=4)$ (a) almost matches the nearly perfect in-context predictive performance of a vanilla transformer, (b) almost perfectly solves in-context path planning problems and (c) learns in-context latent graphs nearly isomorphic to the ground truth.

Finally, our TDB$(S=3, \enc, M=4)$---which also predicts the next latent encoding---performs worse than the other models. This is explained by its large variances across the different experiments. In fact, we observe that for two out of ten experiments, its training accuracy remains very low---below $40\%$---as the model struggles to learn to predict the next observation. For the remaining eight runs, prediction and path planning performance are near optimal and the model competes with our best \texttt{TDB}$(S=3, M=4)$. We could try increasing the number of bottlenecks and training a TDB$(S=3, \enc, M=16)$ to mitigate this issue.

\subsection{In-context accuracy by timestep}
\begin{figure*}[!h]
    \centering
    \includegraphics[width=0.45\textwidth]{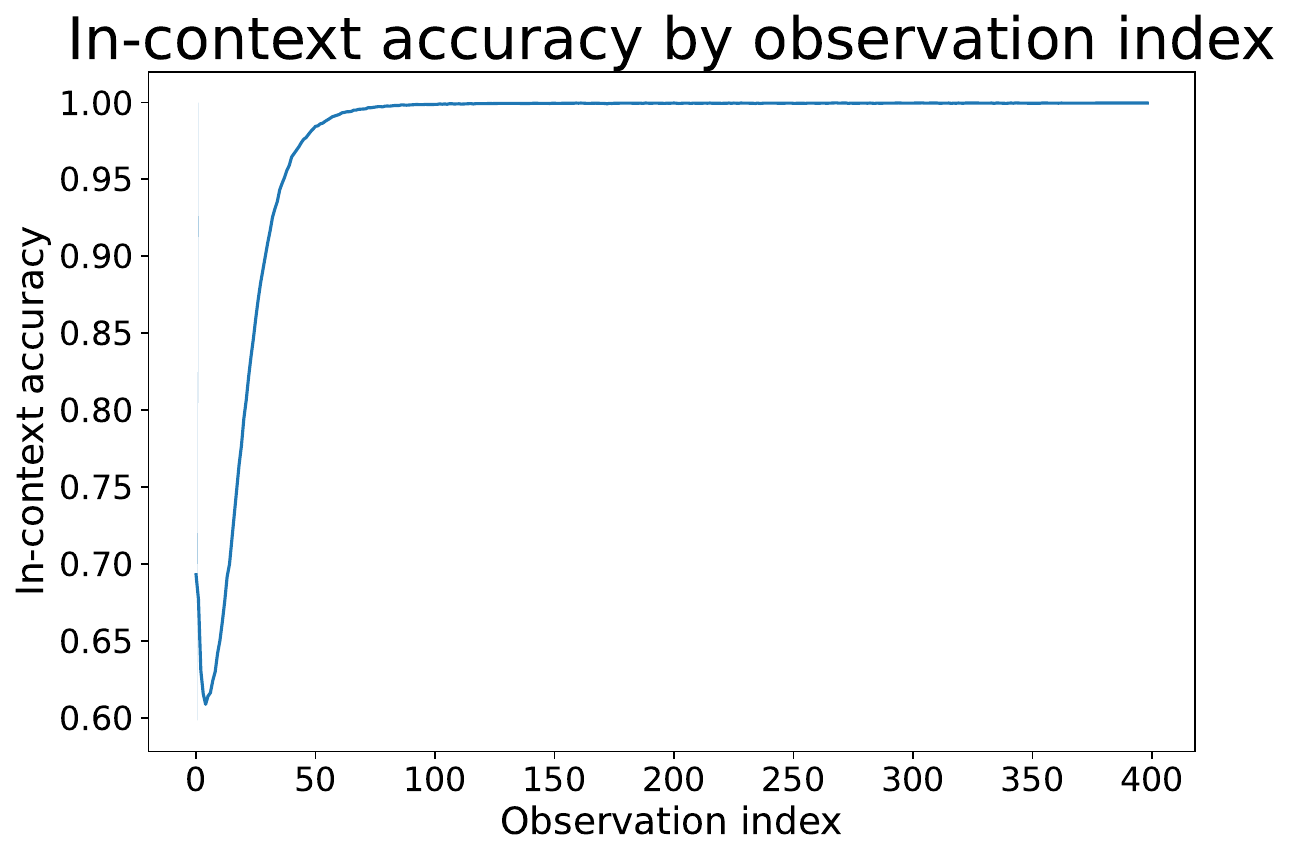}
    \caption{After $70$ timesteps, in-context accuracy for a \texttt{TDB}$(S=3, M=4)$ trained on $5$k training rooms becomes higher than $99\%$.}
    \label{fig:appendix_prediction_accuracy_by_obs_for_icl}
\end{figure*}

Fig.\ref{fig:appendix_prediction_accuracy_by_obs_for_icl} shows the average in-context accuracy on the new test rooms, as a function of the observation index, for our best \texttt{TDB}$(S=3, M=4)$ trained on $5$k rooms. Interestingly, in-context accuracy drops after a few iterations. Indeed, as the model sees new observations in the new test rooms, the mapping between the base labels $\tilde{x}_n$ and the observations $x_n$ may change---which may confuse the model. After a few tenths of iterations, \texttt{TDB} is able to locate itself in the new test room. After $70$ iterations, in-context accuracy becomes higher than $99\%$.

\subsection{In-context learning is driven by spatial exposure to base targets in the training data} \label{appendix:icl_spatial}

\textbf{Restricting training rooms: }
For this experiment, we build training rooms such that the $k$th highest color (almost) never appears at any of the spatial positions indicated by the set $\mathcal{S}_k$ of the room partition. With the notations of Appendix \ref{appendix:icl_setup}, each injective mapping $\phi$ used to generate a training room has to satisfy the rule
$$\mathbf{\mathcal{R}} = \{ \forall k \le O: ~ i_k \ne k \}.$$ 

For instance, the mapping $\phi(1)=5, ~ \phi(2) = 13, ~\phi(3) = 20, ~\phi(4) = 8$ cannot be used to generate a training room. Indeed, when we rank the entries we get $i_1=1, ~i_2=4, ~i_3=2, ~i_4=3$ and $i_1=1$ violates $\mathbf{\mathcal{R}}$. Similarly $\phi(1)=20, ~ \phi(2) = 2, ~\phi(3) = 15, ~\phi(4) = 11$ violates $\mathbf{\mathcal{R}}$ as $i_3=3$.

Given a training sequence $x$, let $n^*$ be such a large enough integer such that all the $O$ different room observations have been seen between $x_1$ and $x_{n^*}$.
For $n\ge n^*$, let $k \le O$ be such that the spatial position $\pos_n$ of the agent at timestep $n$ belongs to the set $\mathcal{S}_k$ of the room partition. Then, by definition, the base target $\tilde{x}_n$ satisfies $\tilde{x}_n \ne k$. Note that, by definition, when $n\le n^*$ and we have not yet observed all the $O$ different room observations, we may have $\tilde{x}_n = k$ when $\pos_n \in \mathcal{S}_k$.

Consequently, for each room spatial position, there is a position-specific base target that the agent will never see (when $n \ge n^*$) at this position on the training set.

We generate $5$k such training rooms and train a \texttt{TDB}$(S=3, M=4)$ as before.

\medskip 

\textbf{In-context accuracies: }
We consider four families of test rooms:
\newline \textbf{(A)}: Test rooms where $\mathbf{\mathcal{R}}$ is respected.
\newline \textbf{(B)}: Test rooms where $\mathbf{\mathcal{R}}$ is violated once.
\newline \textbf{(C)}: Test rooms where $\mathbf{\mathcal{R}}$ is violated twice.
\newline \textbf{(D)}: Test rooms where $\mathbf{\mathcal{R}}$ is violated four times. These are the rooms satisfying $\phi(0) \le \phi(1) \le \phi(2) \le \phi(3)$.

On test rooms \textbf{(A)}, in-context accuracy is $98.05\%(\pm0.07\%)$.
\newline On test rooms \textbf{(B)}, in-context accuracy is $56.97\%(\pm0.82\%)$.
\newline On test rooms \textbf{(C)}, in-context accuracy is $50.25\%(\pm0.51\%)$.
\newline On test rooms \textbf{(D)}, in-context accuracy is $46.20\%(\pm0.73\%)$.

This drastic drop in in-context accuracy demonstrates that---in addition to the number of training rooms---in-context capacities are also driven by spatial exposure to base targets during training.